\pdfoutput=1
\documentclass[11pt]{article}
\usepackage[preprint]{acl}
\usepackage{times}
\usepackage{latexsym}
\usepackage[T1]{fontenc}
\usepackage[utf8]{inputenc}
\usepackage{microtype}
\usepackage{inconsolata}
\usepackage{graphicx}
\usepackage{caption}
\usepackage{multirow}
\usepackage{amsmath}
\usepackage{subcaption}
\usepackage{longtable}
\usepackage[breakable,skins]{tcolorbox}
\usepackage{algorithm}
\usepackage{algorithmic}
\usepackage{tabularx}
\usepackage{array}
\usepackage{newfloat}
\usepackage{listings}
\DeclareCaptionStyle{ruled}{labelfont=normalfont,labelsep=colon,strut=off}
\floatstyle{ruled}
\newfloat{listing}{tb}{lst}{}
\floatname{listing}{Listing}

\usepackage{booktabs}

\usepackage{paralist}

\newcommand{\theTitle}{The Machine's Internal Clock: Do LLMs Share Human Temporal Illusions?}

\title{\theTitle}
\author{Catherine Bao \quad Vivek Srikumar \\
  University of Utah \\
  \texttt{u1459030@utah.edu}, \texttt{svivek@cs.utah.edu}}

\begin{document}

\maketitle

\begin{abstract}
Human perception of time is subjective. Temporal illusions show that the brain relies on context and relational cues when judging duration. Prior studies established these effects with visual and auditory stimuli. We ask whether written narratives alone can cause these illusions, and whether LLMs respond as human participants do. We introduce a benchmark of 6,684 paired scenarios spanning five illusions, each judged from the character's and the reader's perspective. In our human study (60 participants), participants' judgments matched the predicted directions for all five illusions in at least one perspective (7 of 10 conditions). Across 14 LLMs, models reproduce every effect humans show in the same direction, but on average more strongly. Most effects persisted even after cue words were removed and after controlling for 15 lexical and semantic features. These results show that models and humans agree in direction but not in strength, which cautions against treating LLM judgments as substitutes for human responses for subjective temporal reasoning.

\end{abstract}

\section{Introduction}
\label{sec:intro}

\begin{figure}[t]
    \centering
    \includegraphics[width=\columnwidth]{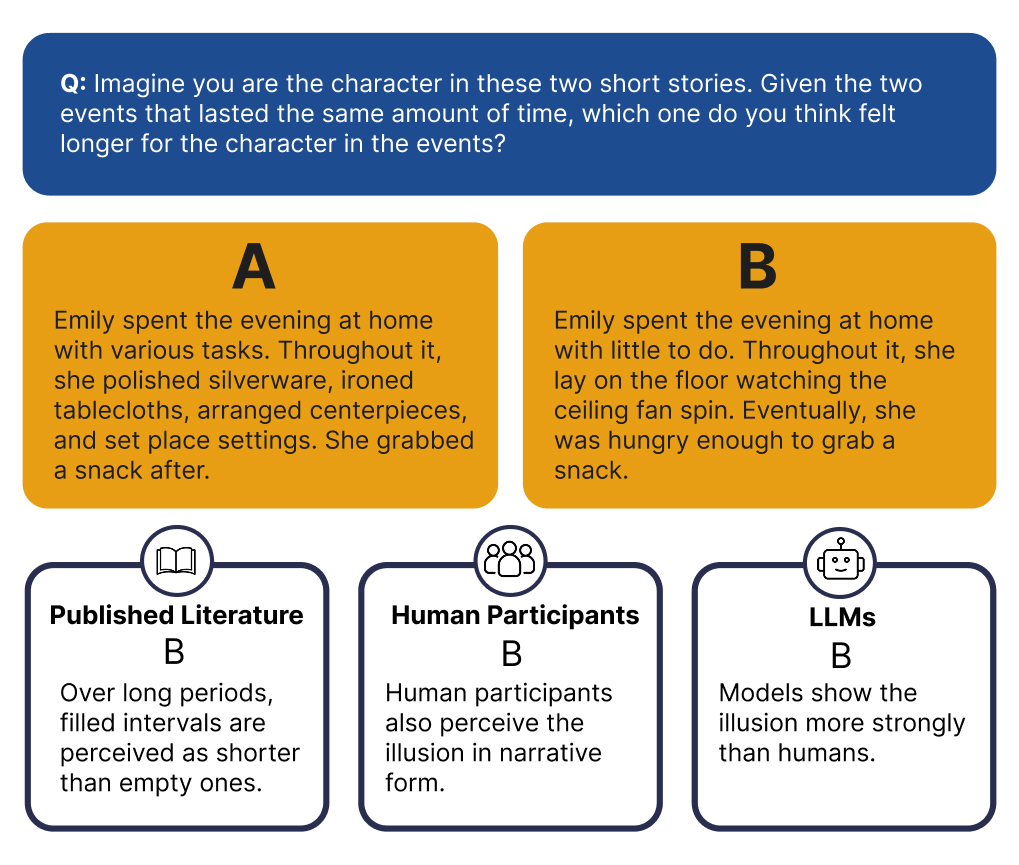}
    \caption{
    We created narrative comparisons to elicit temporal illusions discovered in psychology literature. Human participants choose the literature-aligned options less strongly than the models. The example demonstrates the filled-duration illusion from the character's perspective (gold set: humans 0.70, models 0.84 on average).
    }
    \label{fig:teaser}
\end{figure}

Our perception of time is subjective. Event density, novelty, and association can affect the perceived duration of events.
Alternative frameworks, such as retrospective timing or purely cognitive models, suggest that we construct our sense of time from experience~\cite{boltz1998processing, block2014Timing}.
Temporal illusions, such as emotional time dilation and the oddball effect, show the gap between objective time and its subjective experience.

LLMs are now a standard method to simulate human participants at lower cost. Recent work has shown that these simulations demonstrate the direction of human behavior by successfully recreating social experiments but tend to overestimate their effects~\cite{Cui2025, Ashokkumar2026}. We extend this investigation to perceptual distortions. In our results, a pilot study run on an LLM predicts every illusion humans show, but overestimates the extent to which they occur. This matters for agents that stand in for survey or UX-study respondents (e.g., rating how long a wait felt) and for systems that interpret people's reports of perceived time, since such models may exaggerate effects humans show weakly.

We ask: \emph{do LLM responses reflect the subjective view of time that humans show, and to the same degree?} A participant who reads a passage imagines a situation from their experience rather than actually living through it. A model predicts what would feel longer to a person, rather than reproducing how an internal timing mechanism would behave.

Time-based questions are difficult for LLMs.
Recent work shows that they struggle with both mathematical and relative time-based tasks, scoring between 60-80\% depending on the question format~\cite{su2024timobettertemporalreasoning, fatemi2024testtimebenchmarkevaluating, wang2024trambenchmarkingtemporalreasoning}.
However, these evaluations target objective temporal reasoning. Work on subjective temporal cognition has examined how models represent the distance between points in time~\cite{li2026othermind} or interpret temporal meaning in narrative~\cite{de-langis-etal-2025-llms}, but not how context expands or compresses perceived duration.

To address this gap, we present a new benchmark that uses temporal illusions to study how context affects temporal perception, and we adapt these illusions into narratives to evaluate human participants and LLMs on the same terms. Literature in psychology studies temporal illusions using visual and auditory stimuli rather than textual ones. Our benchmark spans five illusions, translated into 6,684 paired scenarios generated from 115 templates.

We ask our evaluation from both a reader and a character's point of view. The character’s perspective asks participants to project, and the reader’s perspective asks about their own experience of reading. In Figure~\ref{fig:teaser}, Emily spends an evening at home either busy with household tasks or watching the ceiling fan spin~\cite{droit2010time}. From the character’s perspective, the literature predicts the empty evening feels longer (experimental passage). From the reader’s perspective, judged over the few seconds of reading, the filled passage should feel longer (control passage)~\cite{buffardi1971factors, schiffman1977role}. This split also tests whether LLMs can make the same distinction.

We chose text over multimodal models because most people interact with AI through text~\cite{Chatterji2025HowPeopleUse}. Video LLMs still struggle with objective temporal perception~\cite{liu-etal-2024-tempcompass, li2026timeblindspatiotemporalcompositionalitybenchmark}, so a visual stimulus would add a confound we want to avoid. It is also not yet known whether text alone can elicit temporal illusions in humans, and this paper tests that directly. Multimodal evaluation is left for future work.

Using this benchmark, we conduct a human study to investigate whether written narratives alone preserve temporal illusions. Human participants perceive the illusions in 7 of 10 illusion and perspective conditions through text. We also examine whether 14 LLMs produce temporal judgments that align with human behavior and/or published research findings. Models reproduce every effect humans show in the same direction, but on average more strongly: they choose the literature-aligned scenario 0.17 more often than humans (significant in all conditions except emotional time dilation and filled duration from the reader's perspective), and their character--reader gap for filled duration is nearly three times the human gap (+0.40 vs.\ +0.14).

In summary, our contributions are:
\begin{itemize}
    \item We introduce a narrative benchmark based on five temporal illusions for testing whether LLMs' temporal judgments align with previously documented human distortions of time.
    \item In a human study, we find evidence that all five illusions transfer to a purely textual narrative format from at least one perspective.
    \item We find that LLMs reproduce every human-aligned effect in the same direction, generally with larger effect sizes: models can mimic which effects humans show, but not the extent.
\end{itemize}

\section{Background and Related Work}
\label{sec:background}

\subsection{Human Temporal Perception in Psychology}

Psychological research on temporal illusions has studied visual or auditory stimuli over short durations~\cite{tipples2008emotional, gil2012emotional, thomas1974filled, wearden2007internal, buffardi1971factors}. In this work, we focus on five illusions, selected for their reliance on attention, density, novelty, and association~\cite{eagleman2008human}, which make them suitable for a narrative representation. They fall into two categories:
\begin{inparaenum}[(a)]
\item Arousal-based illusions (Emotional Time Dilation and the Oddball Effect), and
\item Information-processing illusions (Filled-Duration, Temporal Order, and Familiarity-Duration).
\end{inparaenum}

\emph{Emotional Time Dilation} arises from physiological arousal and attention; researchers typically study it using visual cues of high-arousal stimuli, such as pictures of phobias or physical mutilation. For example, people with a fear of spiders overestimate how long they spent watching a spider, compared with people without that fear~\cite{watts1984fear}. Figure~\ref{fig:teaser_templates} shows how we carry this into text: Andrew casually checks his email while he notices a flickering reflection in the puddle outside or a clown chasing a child under a streetlight.

Since the distortion itself is tied to arousal and attention, the phenomenon can also extend to non-emotional contexts. The \emph{Oddball Effect} demonstrates this: when an unexpected stimulus appears within a series of repeated expected stimuli, the duration of the unique item seems longer~\cite{pariyadath2007effect}. The novelty of the oddball relative to the repeating standard items drives the effect~\cite{birngruber2015introducing, pariyadath2007effect}. For example, one unexpected photo in a slideshow of near-identical vacation shots seems to stay on screen longer.

Stimulus density and complexity affect illusions based on processing information. Intervals containing discrete elements, such as tones and visual markers, are perceived as longer than empty intervals of the same duration~\cite{thomas1974filled,wearden2007internal}.
This is the \emph{Filled-Duration Effect}. As the discrete components within an interval increase, the subjective duration also increases~\cite{buffardi1971factors, schiffman1977role}. These studies covered short durations. For long periods, when these elements are structured, such as music, time may be thought of as shorter, suggesting that the ``flow'' of organized complexity can compress time over long durations~\cite{droit2010time}. Put plainly, a busy few seconds feels longer than an empty few seconds, but a busy evening feels shorter than an empty one.

Boiling water should feel quicker than assembling a bookshelf for the first time, even when both take the same time (Linda's pair in Appendix Table~\ref{tab:template_examples}). This is the \emph{Familiarity-Duration Effect}: familiar items cause a shorter neural response and subjective time~\cite{manahova2020familiarity, avant1975stimulus, schiffman1977role}, as the brain processes them more efficiently~\cite{skylark2017further}. As a person's physical space becomes more familiar, the mental map of space expands while the estimate of time taken to traverse it contracts~\cite{jafarpour2017familiarity}. \emph{Temporal Order Effects} prioritize the sequence of events over the durations. As the brain diverts resources, the mental effort required to process sequence events decreases the accuracy of duration judgment~\cite{brown2014time}. For example, someone who realizes they are tying their shoelaces in the wrong order has to track each step, and the task seems to take longer.

\subsection{Existing LLM Benchmarks}

Recent work has drawn from cognitive psychology to study LLM behavior~\cite{doi:10.1073/pnas.2218523120}. Especially relevant is the line of work on theory of mind, which examines whether models can reason about beliefs, perspectives, and mental states~\cite{doi:10.1073/pnas.2405460121, nickel2024probingrobustnesstheorymind}.

Existing temporal benchmarks are either factual, mathematical, or reasoning-based. Factual and math-based benchmarks ask models to look up timestamped knowledge~\cite{chen2021datasetansweringtimesensitivequestions, tan2023benchmarkingimprovingtemporalreasoning, fatemi2024testtimebenchmarkevaluating, wei-etal-2023-menatqa} or to do arithmetic within calendar or clock systems~\cite{su2024timobettertemporalreasoning}, which measures a model's retention and not its perception of duration.

Reasoning-based temporal questions require models to infer the frequency, duration, or plausibility of events from their temporal context. They test ``temporal common sense'' and plausible causal relationships~\cite{zhou2019goingvacationtakeslonger, su2024timobettertemporalreasoning, wang2024trambenchmarkingtemporalreasoning}. These ask for an objective duration estimate in response to subjective questions (e.g., `The chairman said the deal remains of substantial benefit. How long did the chairman speak?'), or ask the model to understand likely social constraints (e.g., why a character like Amy would choose to `start laundry early in the morning every weekend')~\cite{su2024timobettertemporalreasoning}.

The most closely related work on subjective measures of time is by~\citet{chen-etal-2025-perceive}, who evaluate temporal relativity in LLMs. Unlike other benchmarks, the study recognizes the relationship between time perception and context using probes such as `How long does a stressful event feel?'.
This moves the evaluation toward subjective judgments.
By providing one objective answer to a subjective question, the study overlooks that perceived duration can change depending on context. For example, an hour-long exam may feel longer to an unprepared student than to one who is well prepared and engaged.

All these datasets assume a real-world ground truth, but many subjective, time-based questions lack fixed answers. Comparisons between experiences, rather than absolute values, shape our perception of time.
Instead, we ask for comparisons of two objectively equal durations. The psychology literature, which studies how temporal distortions are triggered and measured, establishes our ground truth.
Our human evaluation addresses whether written narratives can elicit behavior aligned with experiencing temporal illusions. Our LLM evaluation determines whether models align with the human time distortions discussed in existing literature and human participants.

\section{Benchmark Design}
\label{sec:benchmark_design}

\subsection{Design Principles}
\label{sec:design_principles}

Our benchmark translates psychology experiments into narrative prompts. Each template consists of a paired control and experimental condition that differ in a targeted manipulation. The experimental condition is the scenario that, based on prior psychological literature, is expected to be perceived as longer for the character. For Filled Duration, the literature predicts the inverse from the reader's perspective, so the control (filled) condition is the expected result (Appendix~\ref{sec:benchmark_design_full}). We therefore report the literature-aligned rate: \(P(\text{experimental})\) in every condition except Filled Duration from the reader's perspective, where it is \(P(\text{control}) = 1 - P(\text{experimental})\). Values above 0.5 indicate the predicted illusion.

Since text length could affect perceived duration, we write the paired conditions to match closely in length and narrative structure. Where the question design allows, templates include both longer and shorter versions to test for this effect.
We ask participants to judge duration both from the character's perspective (lived time) and the reader's perspective (observed/processed time).

Appendix~\ref{sec:benchmark_design_full} provides full details of the design process for each illusion (prompt templates and examples). Here we describe one illusion.

\subsubsection {Arousal-Based Illusions: A Worked Example}

We use Emotional Time Dilation as a running example of our design process, including how we distinguish it from the closely related Oddball Effect below.

The difference between emotional time dilation and the oddball effect is what captures attention and how strongly.
Heightened arousal (events that feel important, stressful, or unexpected) drives emotional time dilation. These situations create top-down attention, meaning internal states such as concern, urgency, or significance guide attention. Our design of the oddball effect is bottom-up novelty. A stimulus stands out within a predictable sequence and briefly captures attention because it is different, not because it carries emotional weight or consequence.

Within this framing, we define emotional time dilation via three mechanisms that cause sustained top-down attention:

\begin{itemize}
    \item[A] \textit{Stakes and Expectations}: we present a neutral stimulus (e.g., a flickering reflection) in both low-stakes and high-stakes contexts, testing how potential consequences alter engagement with otherwise identical stimuli.
    \item[B] \textit{Interval Intensity}: we vary cognitive demand in an empty interval by comparing high-focus tasks with low-demand activities. When cognitive resources are heavily engaged, fewer remain available for tracking time, altering how long the interval feels~\cite{brown2014time}.
    \item[C] \textit{Emotional Arousal}: we pair a neutral baseline with a sudden or emotional event while holding context and stakes constant, testing whether involuntary attentional capture expands perceived duration.
\end{itemize}

By holding the objective action constant while varying the emotional context, we test whether participants perceive the same event as lasting longer when the scene feels more intense. The remaining four illusions follow the same pattern: the mechanism from the source literature determines the subtypes, and each subtype becomes a matched, length-controlled pair. 

\begin{figure}[htpb]
    \centering
    \includegraphics[width=\columnwidth]{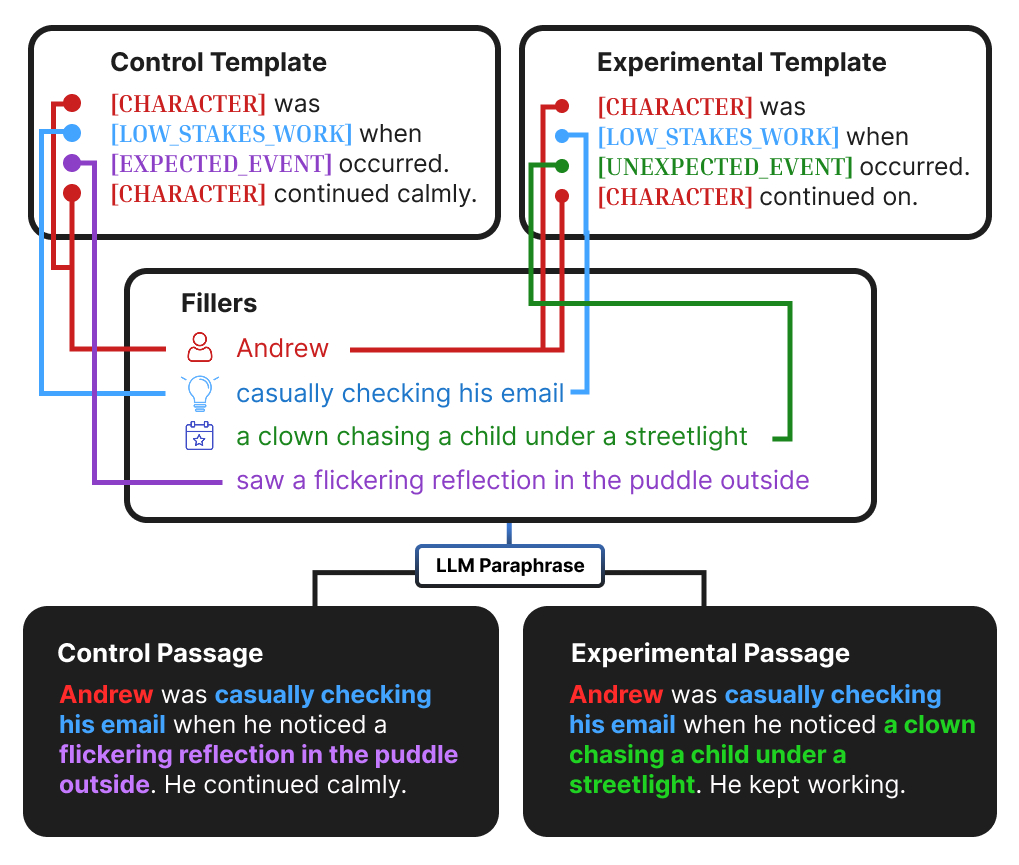}
    \caption{
    This example shows a template and instantiation of a pair of the Emotional Arousal subcategory of Emotional Time Dilation.
    }
    \label{fig:teaser_templates}
\end{figure}

\subsection{Dataset Creation}
\label{sec:dataset_creation}

Using the design principles for each illusion type, we create and populate slot-based templates with fillers using a Jinja-based generation framework.\footnote{\url{https://github.com/utahnlp/madlibs}}
We design our fillers (stimuli) for contextual variation while preserving the structure of each illusion type. Refer to Figure~\ref{fig:teaser_templates} as an example.
We wrote the initial set of fillers per category manually, expanded it using GPT-4o~\cite{openai_gpt4o_system_card}, and then reviewed and edited the set by hand. List sizes depend on what each slot must hold (3--100 entries; Appendix~\ref{appendix:Appendix_dataset}). We sampled 50 character names at random from the top 100 most common names in the United States~\cite{ssa_babynames_century}. We generated 2000 paired scenarios per illusion type by randomly sampling from the available templates, for 10000 candidate paired scenarios. We also manually created a gold set of 250 items (50 per illusion type), edited by hand for clarity and representation of each illusion.

To reduce stylistic noise and improve grammar, we paraphrased each paired scenario three times using Qwen3-32B~\cite{qwen3technicalreport}, few-shot prompted with gold items (Appendix~\ref{appendix:Appendix_dataset}). From these variants, we selected the version that most closely matched in length between control and experimental conditions, minimizing word-count differences that could bias duration judgments. We removed a combination when every paraphrased passage differed in word count from its source passage by more than seven words, leaving 6,684 paired scenarios (Table~\ref{tab:dataset_sizes}, Appendix). Among the remaining variants, we selected the one whose control and experimental passages were closest in word count, with the median difference being 2 words.

While the dataset is not perfectly balanced after filtering, each category retains a large enough sample size for analysis (Table~\ref{tab:subtype_breakdown}).
Because we generate items from templates, examples within a template family are not statistically independent, so the effective sample size is smaller than the raw counts. We will release the complete dataset, slot fillers, anonymized human responses, and associated scripts at \url{https://github.com/utahnlp/temporal-illusions}.

\section{Experimental Setup}
\label{sec:experiments}

We examine whether behavior aligns with human intuition of narrative time perception. Specifically, we ask:
\begin{itemize}
    \item[] \textbf{RQ1:}  Do humans perceive differences consistent with known time dilation effects across narrative templates? 
    \item[] \textbf{RQ2:}  Do language models match human perception and/or replicate known temporal illusions in psychology literature? If so, do they show stronger alignment with some illusions than others?
    \item[] \textbf{RQ3:} Do models distinguish between the character's and reader's perspectives in time perception?
\end{itemize}

We address these questions via two evaluations, described below.

\subsection{Human Evaluation}

We recruited participants for the human evaluation using Prolific~\cite{palan2018prolific} (eligibility and demographics in Appendix~\ref{sec:human_vis}). We compensated participants \$10.04 per hour (above the U.S. federal minimum wage). We recruited 60 participants completing 25 questions in $\sim 8$ minutes. We sampled questions from the gold set of 250 hand-curated items.

We instructed participants to rely on their first intuition and respond quickly. For each question, participants evaluated randomized scenarios from one of two perspectives. The first, the character’s perspective, asked: \textbf{“Which event felt longer for the character experiencing it?”} The second, the reader’s perspective, asked: \textbf{“Which event felt longer to read and process?”}. The University of Utah Institutional Review Board reviewed and approved this study. All participants provided informed consent before beginning the study and could withdraw at any time without penalty. For additional details on survey instructions, see Appendix~\ref{sec:eval_instructions}.

\subsection{LLM Evaluation}
\label{sec:llm_eval}

\begin{table}[t]
    \centering
    \small
    \setlength{\tabcolsep}{4pt}
    \begin{tabular}{ll}
    \toprule
    \textbf{Model} & \textbf{Sizes} \\
    \midrule
    \multicolumn{2}{c}{Open models}\\
    Gemma 3~\cite{gemma_2025} & 4B, 12B, 27B \\
    gpt-oss~\cite{openai2025gptoss120bgptoss20bmodel} & 20B, 120B \\
    Llama 3~\cite{grattafiori2024llama3herdmodels} & 8B, 70B \\
    Qwen3~\cite{qwen3technicalreport} & 8B, 32B \\
    \midrule
    \multicolumn{2}{c}{Proprietary models} \\
    GPT-5.4 mini, GPT-5.4~\cite{openai_gpt54_thinking_system_card} & -- \\
    GPT-5.5~\cite{openai_gpt55_system_card} & -- \\
    Claude Sonnet 4.6~\cite{anthropic_sonnet_46_system_card} & -- \\
    Claude Haiku 4.5~\cite{anthropic_haiku_45_system_card} & -- \\

    \bottomrule
    \end{tabular}
    \caption{LLMs used in our evaluation. Proprietary model sizes are undisclosed.}
    \label{tab:models}
\end{table}

We evaluated a diverse set of models (Table~\ref{tab:models}) on both the curated gold set and a broader filtered dataset (decoding settings in Appendix~\ref{sec:config}; positional bias, prompt complexity, and non-committal responses in Appendix~\ref{sec:additional-results}).
To keep consistency with the human evaluation, we prompt each model using the same instructions given to participants (Appendix~\ref{sec:evaluation_prompts}).
We conduct both a 2-way evaluation, in which the model selects between two randomized scenarios as shown in Figure~\ref{fig:teaser}, and a 4-way evaluation, which additionally allows responses of \emph{same} or \emph{cannot tell}.
In the former, models choose between Scenario A and Scenario B. The latter adds \emph{same} and \emph{cannot tell} as options to help measure decisiveness and confidence.

\subsubsection{Lexical and Semantic Confound Analysis} To test whether model choices could be explained by surface differences between the passages rather than by temporal judgment, we ran a lexical and semantic confound analysis for both the human and the LLM evaluations (Appendix~\ref{sec:confound_details}).

\subsubsection{Cue-Neutralization Test} Each gold item was rewritten twice by an LLM (Gemma 3 27B). One author manually checked the rewritten pairs, and 95\% preserved the original events and order. In the cue-neutralized version, words that name or evaluate the manipulation were removed, while the characters, events, and order stayed the same. In the paraphrase version, the question set was reworded with the original paraphrasing prompt, so differences reflect cue removal rather than rewording (Appendix~\ref{sec:confound_details}).

\section{Results}
\label{sec:results}

In this section, we report the human evaluation results (\S~\ref{sec:results-human}, addressing \textbf{RQ1}), followed by the LLM evaluation (\S~\ref{sec:results-llm}, addressing \textbf{RQ2}, \textbf{RQ3}). Responses can depend on the order in which we present each scenario, so we correct positional bias by symmetrizing results across the two orderings (experimental first vs.\ second), following~\citet{li-etal-2020-unqovering}.

\subsection{Human Evaluation}
\label{sec:results-human}

We used the \(2\)-way evaluation to make weak preferences more observable in aggregated statistics.

Human evaluation showed significant effects in \(7\) of the \(10\) illusion--perspective combinations, suggesting that written narratives alone can elicit temporal illusions (\textbf{RQ1}, Figure~\ref{fig:human_vs_model}; human-only results in Appendix Figure~\ref{fig:human_illusions_results}). The three non-significant combinations were emotional time dilation and filled duration from the reader's perspective, and temporal order from the character's perspective. Filled duration from the reader's perspective is the one condition where the literature predicts the control (filled) interval: models lean toward the filled interval there (0.41 on the gold set, 0.24 on the full dataset), while human participants show no reliable preference. The character--reader difference seen in LLMs also appears in humans for filled duration: humans showed a larger filled-duration effect from the character's perspective than from the reader's perspective.

\begin{figure*}[!t]
    \centering
    \includegraphics[width=0.75\textwidth]{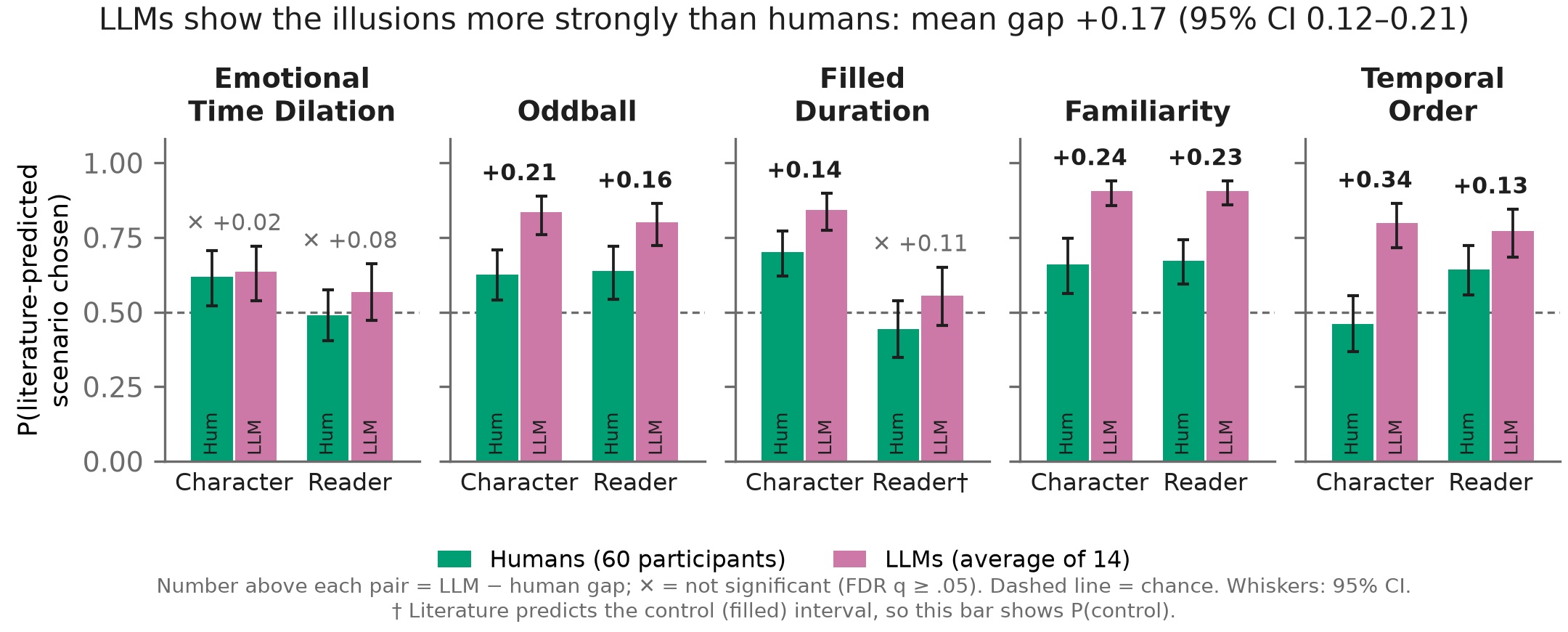}
    \caption{
    Humans vs.\ LLMs on the gold set from a joint mixed-effects model of human and LLM trials.
    }
    \label{fig:human_vs_model}
\end{figure*}

Where humans show these illusions, models show them more strongly on average (Figure~\ref{fig:human_vs_model}). In a joint mixed-effects model of human and LLM trials on the gold set, models chose the literature-aligned scenario \(0.17\) more often than humans on average (95\% CI 0.12--0.21; Appendix~\ref{sec:stats}), and in the same direction wherever humans showed an illusion. 
The model--human gap was significant for oddball, familiarity, and temporal order in both perspectives, and for filled duration from the character's perspective, but not for emotional time dilation or filled duration from the reader's perspective (where the literature predicts the control interval, so the gap is measured toward the control). For filled duration, both groups showed the expected perspective reversal, but it was larger for models.

We fitted the same confound analysis to the human responses. No surface feature reliably predicted human choices (all $q \ge 0.43$), and the human familiarity and oddball point estimates were essentially unchanged by the controls. Surface features therefore do not account for the human familiarity and oddball effects. The controls do widen the human intervals, so two character-perspective effects are less robust: filled duration drops from 0.70 to 0.63, and oddball (0.61 to 0.60) is no longer significant (Figure~\ref{fig:residual_analysis}). Humans leaned slightly toward passages with more psychology cue words in the same direction as the models (\(\mathrm{OR} = 1.18\) vs.\ \(1.24\)), with no significant difference between the two (ratio 0.94, 95\% CI 0.67--1.32).

\subsection{LLM Evaluation}
\label{sec:results-llm}

We first validate that the paraphrased dataset reproduces the hand-curated gold set. Agreement between the two is high for the character's perspective (Spearman rank correlation $\rho = 0.79/0.85$ for 2-way/4-way, respectively) and the reader's perspective (Spearman rank correlation $\rho = 0.86/0.84$ for 2-way/4-way, respectively). In general, both the 2-way and 4-way evaluation have almost equivalent analysis results.
Appendix~\ref{appendix:Appendix_eval} includes
the full concordance, rank-preservation and detailed model-specific results for both evaluations.

\subsubsection{Differences across Illusion Types}

\begin{figure}[t]
    \centering
    \includegraphics[width=0.88\columnwidth]{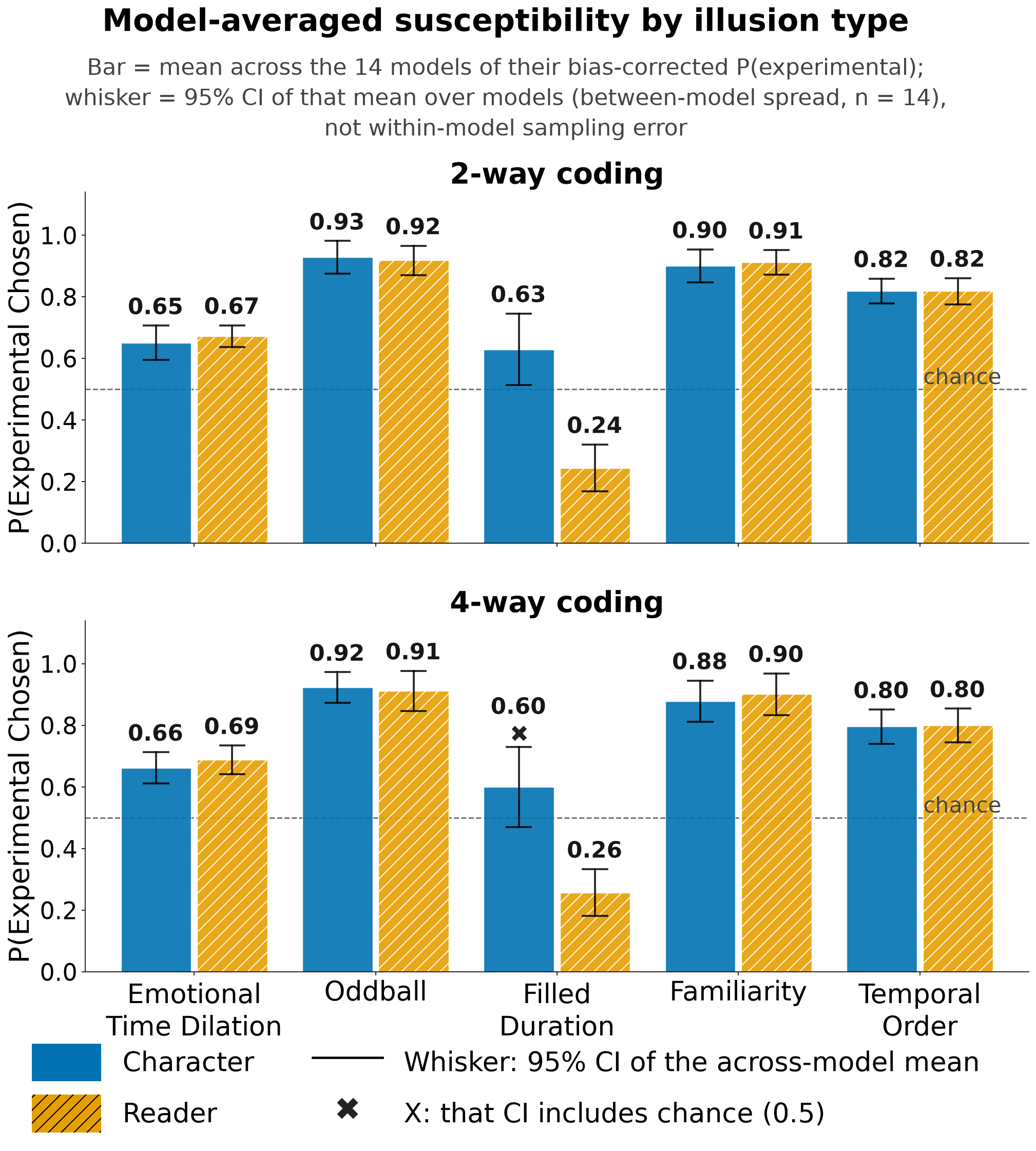}
    \caption{P(Experimental Chosen) by illusion and perspective, averaged over 14 models.}
    \label{fig:mlr_aggregate}
\end{figure}

Illusion strength varies across the five illusion types (\textbf{RQ2}, Figure~\ref{fig:mlr_aggregate}).
The bars report the proportion $p_{\text{exp}}$ of 2-way responses in which models select the experimental scenario, with whiskers showing 95\% CI of the cross-model mean.
The size of the departure from chance is more important than its existence, so we read effects off the intervals rather than significance stars.

Oddball and Familiarity reach high $p_{\text{exp}}$ across nearly all models (0.85--1.00, confidence intervals well clear of chance), and Temporal Order is close behind (0.82). Emotional Time Dilation is much weaker. Both Emotional Time Dilation and Oddball are arousal-related illusions, but Emotional Time Dilation uses emotionally salient contexts (e.g., surprise or stress) while Oddball relies on more structured and systematic changes. Filled Duration depends on perspective (discussed below).
\subsubsection{Character's vs.\ Reader's Perspective}

There are clear differences between the character's and the reader's perspectives (Figure~\ref{fig:mlr_aggregate}; per model in Figures~\ref{table:reader_character_4way} and~\ref{fig:reader_character_2way}, \textbf{RQ3}) for Filled Duration: the model's selection changes from the experimental option when moving from the character's to the reader's perspective. Prior work suggests that filled intervals are perceived as longer at short durations, and empty intervals are perceived as longer at longer durations. Our results are consistent with this distinction (Appendix~\ref{sec:benchmark_design_full}). 

From the character's perspective, where the narrative frames the experience as lived through, models favor the empty interval as longer. The reader's perspective, judged over the short time it takes to read the narrative, favors the filled interval, as the short-interval studies predict.

Other illusion types show smaller shifts between perspectives. Rather than reflecting ``correct'' or ``incorrect'' responses, these differences indicate that some illusions depend on how the narrative frames the temporal experience. An analysis by subtype is provided in Appendix~\ref{sec:subtypes}.

\subsubsection{Thinking Tokens}

To probe the strong alignment with the literature, we analyzed the thinking traces of Qwen3-32B, one of the largest evaluated models with exposed thinking tokens. Although the prompt never asks for a psychological explanation, 38.6\% of its gold-set traces explicitly appeal to psychological research or a named theory (e.g., `cognitive load theory'). These research-framed traces choose the experimental option at about the same rate as other traces (80.7\% vs.\ 81.7\%), so the framing is unlikely to explain why models show the illusions more strongly than humans (Appendix~\ref{sec:trace_patterns}).

\paragraph{Lexical and semantic confound analysis.}
Surface differences between the two passages do not explain the models' preferences (Figure~\ref{fig:residual_analysis}). Controlling for all 15 measures in 10 feature groups shifts P(exp) by at most 0.06 for models, and the preference patterns remain largely unchanged. The one notable model-level exception is cue words.

\begin{figure}[t]
    \centering
    \includegraphics[width=0.8\columnwidth]{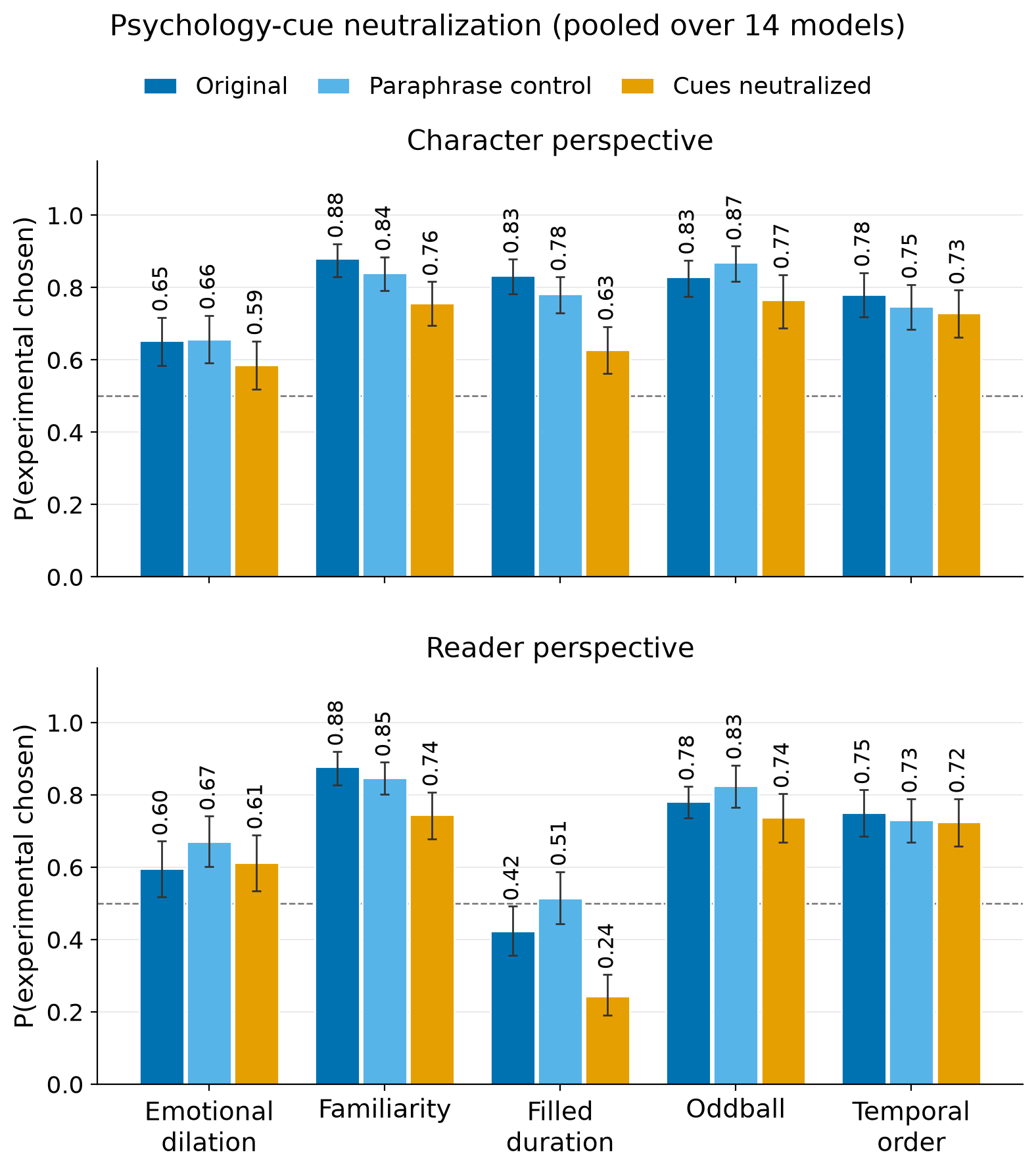}

    \caption{
    Removing cue words leaves most effects unchanged; familiarity drops, and filled duration changes the most. Computed on the 225 gold items whose rewrites passed the automatic checks, so original values differ slightly from Figure~\ref{fig:residual_analysis}.
    \label{fig:cue_ablation}
    }
\end{figure}

\paragraph{Cue-neutralization test.}
Removing cue words leaves emotional time dilation, oddball, and temporal order largely unchanged (Figure~\ref{fig:cue_ablation}). Familiarity drops but stays well above chance, so cue words carry part of this effect. Filled duration changes the most, but in opposite directions: from the character's perspective, the preference for the empty interval weakens toward chance. From the reader's perspective, models shift further toward the filled interval after cue removal, which moves them closer to the literature's prediction. The paraphrase control stays close to the original for most conditions, so these changes come from removing the cue words rather than from rewording. Because removing cue words can weaken the manipulation itself, a drop does not mean the effect was fake, but an effect that survives cannot come from models simply matching on cue words.

\section{Discussion and Future Work}
\label{sec:discussion}

We ask whether LLM responses match the human illusion pattern. We do not take a position on whether they should. Our results show that models and humans agree in direction but differ in strength.

Understanding how models interpret and represent temporal illusions can inform the design of agents. As \S~\ref{sec:intro} notes, an LLM standing in for survey or UX-study respondents would get the direction of these illusions right but overstate their size.

For filled duration, the predicted answer reverses between the character's and the reader's perspective, and both humans and models follow this pattern. This creates a blind spot in temporal-reasoning benchmarks that treat duration as having one correct answer, even though additional context such as point of view and experience can flip answers. Participants showed illusions from solely text-based stimuli. Narrative stimuli are cheap and scalable, making them a practical tool for time-perception research. In NLP research, this also validates using narrative text to study judgment that depends on framing. 

A question remains: why do models show these effects more strongly than humans? One possibility is that models draw on literature, though research-framed traces chose the experimental option no more often than other traces. Traces may be post-hoc explanations. These intuitions may come from patterns learned in training data. It is also possible that noise in human responses weakens the human effects. Another possible explanation is aggregation. As the question design was based on a generalized behavior, it reflects the majority answer rather than acting as a sampled person. This aligns with existing work showing that LLM simulations flatten variation~\cite{Wang2025LargeLM} and with~\citet{de-langis-etal-2025-llms}, who found that LLMs over-rely on prototypical narrative interpretations. Selecting the textbook direction more consistently than human participants may reflect a related effect.

\section{Conclusion}
\label{sec:conclusion}

We introduced a narrative benchmark of 6,684 paired scenarios, generated from 115 templates across five temporal illusions, to ask whether written text can elicit illusions. Our human study finds that they transfer in 7 of 10 illusion–perspective conditions, though with limited power. On average, models select the literature-predicted scenario in the same direction as humans but more strongly.

\section*{Limitations}
\label{sec:limitations}

While we filtered for word count differences, some still exist across illusion types. This is unavoidable because certain illusion templates require longer descriptions structurally. Relatedly, some illusions, e.g., Familiarity, depend on assumptions about experiences that are not universal. A task that is routine for one participant may be novel for another. All items are in English and reflect U.S.-centric names and everyday activities, and our human participants were U.S. residents whose primary language is English, so our findings may not hold for other languages or cultures.

Our narrative task asks participants to judge described intervals rather than experience timed stimuli, so responses may reflect beliefs about time as well as perceptual distortion. Future work could test whether participants familiar with the psychology literature answer differently.

Additionally, humans answered on first intuition, while several models reasoned first to varying degrees. Qwen3 and gpt-oss produce visible thinking, and Claude ran with thinking off. GPT-5.4 mini, GPT-5.4, and GPT-5.5 ran at their API default reasoning.

We also observe substantial variation in positional bias across models. On the full dataset, order is randomized once per pair, so each ordering covers a different subset of items (gold items are run in both orders). We mitigated it with UNQOVER-style symmetrization but did not completely eliminate it. Evaluation results may reflect model-specific artifacts rather than a meaningful signal.

A limitation of this work is that we only study the model outputs and reasoning traces, not the internal representations. To better test the underlying mechanisms responsible for these behaviors, future work would require probing the model's internal representations during decision-making. Isolating activation vectors associated with emotion, attention, novelty, or other psychological factors would allow for a more direct comparison to human psychology experiments. 

Our other analyses have limits as well. Reasoning traces are not guaranteed to reflect the computation that produces the final answer, so the $\sim$39\% figure describes how models explain their choices, not necessarily how they make them. Our human study collected relatively few judgments per item (about eight per item shown, or about five per item and perspective), which limits statistical power for item-level analysis. The three non-significant human conditions may reflect limited power rather than absence of an effect. Finally, for some illusions the cue words are part of the manipulation itself, so removing them can weaken the illusion along with the cue. The cue-neutralization test is a check rather than a raw test of hidden cues.

\section*{Acknowledgments}
\label{sec:acknowledgments}

This work was supported by the Parent Fund Scholarship through the University of Utah Undergraduate Research Opportunity Program. The support and resources from the Center for High Performance Computing at the University of Utah are gratefully acknowledged.

\bibliography{cited}

@article{tipples2008emotional,
  title   = {Negative emotionality influences the effects of emotion on time perception},
  author  = {Tipples, Jason},
  journal = {Emotion},
  volume  = {8},
  number  = {1},
  pages   = {127--131},
  year    = {2008},
  doi     = {10.1037/1528-3542.8.1.127}
}

@article{watts1984fear,
  title   = {Fear and time estimation},
  author  = {Watts, F. N. and Sharrock, R.},
  journal = {Perceptual and Motor Skills},
  volume  = {59},
  number  = {2},
  pages   = {597--598},
  year    = {1984},
  doi     = {10.2466/pms.1984.59.2.597}
}

@article{Cui2025,
  author    = {Ziyan Cui and Ning Li and Huaikang Zhou},
  title     = {A large-scale replication of scenario-based experiments in psychology and management using large language models},
  journal   = {Nature Computational Science},
  volume    = {5},
  number    = {8},
  pages     = {627--634},
  year      = {2025},
  doi       = {10.1038/s43588-025-00840-7},
  url       = {https://www.nature.com/articles/s43588-025-00840-7}
}

@article{Wang2025LargeLM,
  title     = {Large language models that replace human participants can harmfully misportray and flatten identity groups},
  author    = {Angelina Wang and Jamie Morgenstern and John Dickerson},
  journal   = {Nature Machine Intelligence},
  volume    = {7},
  pages     = {400--411},
  year      = {2025},
  doi       = {10.1038/s42256-025-00986-z},
  url       = {https://api.semanticscholar.org/CorpusID:276443240}
}

@article{Ashokkumar2026,
  author    = {Ashokkumar, Ashwini and Hewitt, Luke and Ghezae, Isaias and Willer, Robb},
  title     = {Large language models can predict the results of social science experiments},
  journal   = {Nature},
  volume    = {656},
  number    = {8126},
  pages     = {115--122},
  year      = {2026},
  month     = {August},
  doi       = {10.1038/s41586-026-10742-x},
  url       = {https://www.nature.com/articles/s41586-026-10742-x}
}

@article{Brysbaert2014,
  author  = {Brysbaert, Marc and Warriner, Amy Beth and Kuperman, Victor},
  title   = {Concreteness ratings for 40 thousand generally known {English} word lemmas},
  journal = {Behavior Research Methods},
  volume  = {46},
  number  = {3},
  pages   = {904--911},
  year    = {2014},
  doi     = {10.3758/s13428-013-0403-5}
}

@inproceedings{HuttoGilbert2014,
  author    = {Hutto, Clayton J. and Gilbert, Eric E.},
  title     = {{VADER}: A Parsimonious Rule-Based Model for Sentiment Analysis of Social Media Text},
  booktitle = {Proceedings of the International AAAI Conference on Web and Social Media},
  volume    = {8},
  pages     = {216--225},
  year      = {2014},
  doi       = {10.1609/icwsm.v8i1.14550}
}

@inproceedings{mohammad-2018-obtaining,
  author    = {Mohammad, Saif},
  title     = {Obtaining Reliable Human Ratings of Valence, Arousal, and Dominance for 20,000 {English} Words},
  booktitle = {Proceedings of the 56th Annual Meeting of the Association for Computational Linguistics (Volume 1: Long Papers)},
  month     = {jul},
  year      = {2018},
  address   = {Melbourne, Australia},
  publisher = {Association for Computational Linguistics},
  url       = {https://aclanthology.org/P18-1017},
  doi       = {10.18653/v1/P18-1017},
  pages     = {174--184}
}

@ARTICLE{block2014Timing,
  AUTHOR={Block, Richard A.  and Grondin, Simon },
  TITLE={Timing and time perception: A selective review and commentary on recent reviews},
  JOURNAL={Frontiers in Psychology},
  VOLUME={5},
  PAGES={648},
  YEAR={2014},
  URL={https://www.frontiersin.org/journals/psychology/articles/10.3389/fpsyg.2014.00648},
  DOI={10.3389/fpsyg.2014.00648},
  ISSN={1664-1078},  
}

@misc{ssa_babynames_century,
  author       = {{U.S. Social Security Administration}},
  title        = {Top Names Over the Last 100 Years},
  year         = {2024},
  howpublished = {\url{https://www.ssa.gov/oact/babynames/decades/century.html}},
  note         = {Accessed April 2026}
}

@misc{nickel2024probingrobustnesstheorymind,
  title={Probing the Robustness of Theory of Mind in Large Language Models},
  author={Christian Nickel and Laura Schrewe and Lucie Flek},
  year={2024},
  eprint={2410.06271},
  archivePrefix={arXiv},
  primaryClass={cs.CL},
  url={https://arxiv.org/abs/2410.06271},
}

@article{
doi:10.1073/pnas.2405460121,
author = {Michal Kosinski },
title = {Evaluating large language models in theory of mind tasks},
journal = {Proceedings of the National Academy of Sciences},
volume = {121},
number = {45},
pages = {e2405460121},
year = {2024},
doi = {10.1073/pnas.2405460121},
URL = {https://www.pnas.org/doi/abs/10.1073/pnas.2405460121},
eprint = {https://www.pnas.org/doi/pdf/10.1073/pnas.2405460121}}

@article{
doi:10.1073/pnas.2218523120,
author = {Marcel Binz  and Eric Schulz },
title = {Using cognitive psychology to understand {GPT-3}},
journal = {Proceedings of the National Academy of Sciences},
volume = {120},
number = {6},
pages = {e2218523120},
year = {2023},
doi = {10.1073/pnas.2218523120},
URL = {https://www.pnas.org/doi/abs/10.1073/pnas.2218523120},
eprint = {https://www.pnas.org/doi/pdf/10.1073/pnas.2218523120}
}

@article{gil2012emotional,
  title={Emotional time distortions: the fundamental role of arousal},
  author={Gil, Sandrine and Droit-Volet, Sylvie},
  journal={Cognition \& emotion},
  volume={26},
  number={5},
  pages={847--862},
  year={2012},
  publisher={Taylor \& Francis}
}

@article{thomas1974filled,
  title   = {Time perception and the filled-duration illusion},
  author  = {Thomas, E. C. and Brown, I.},
  journal = {Perception \& Psychophysics},
  volume  = {16},
  number  = {},
  pages   = {449--458},
  year    = {1974},
  doi     = {10.3758/BF03198571}
}

@article{droit2010time,
  title={Time flies with music whatever its emotional valence},
  author={Droit-Volet, Sylvie and Bigand, Emmanuel and Ramos, Danilo and Bueno, Jos{\'e} Lino Oliveira},
  journal={Acta psychologica},
  volume={135},
  number={2},
  pages={226--232},
  year={2010},
  publisher={Elsevier}
}

@article{wearden2007internal,
  title={Internal clock processes and the filled-duration illusion.},
  author={Wearden, John H and Norton, Roger and Martin, Simon and Montford-Bebb, Oliver},
  journal={Journal of Experimental Psychology: Human Perception and Performance},
  volume={33},
  number={3},
  pages={716--729},
  year={2007},
  publisher={American Psychological Association}
}

@article{buffardi1971factors,
  title={Factors affecting the filled-duration illusion in the auditory, tactual, and visual modalities},
  author={Buffardi, Louis},
  journal={Perception \& Psychophysics},
  volume={10},
  number={4},
  pages={292--294},
  year={1971},
  publisher={Springer}
}

@article{hurlemann2005noradrenergic,
  title={Noradrenergic modulation of emotion-induced forgetting and remembering},
  author={Hurlemann, Ren{\'e} and Hawellek, Barbara and Matusch, Andreas and Kolsch, Heike and Wollersen, Heike and Madea, Burkhard and Vogeley, Kai and Maier, Wolfgang and Dolan, Raymond J},
  journal={Journal of Neuroscience},
  volume={25},
  number={27},
  pages={6343--6349},
  year={2005},
  publisher={Society for Neuroscience}
}

@article{brown2014time,
  title={Time perception and temporal order memory},
  author={Brown, Scott W and Smith-Petersen, G Andrew},
  journal={Acta psychologica},
  volume={148},
  pages={173--180},
  year={2014},
  publisher={Elsevier}
}

@article{manahova2020familiarity,
  title={Familiarity increases processing speed in the visual system},
  author={Manahova, Mariya E and Spaak, Eelke and de Lange, Floris P},
  journal={Journal of cognitive neuroscience},
  volume={32},
  number={4},
  pages={722--733},
  year={2020},
  publisher={MIT Press One Rogers Street, Cambridge, MA 02142-1209, USA journals-info~…}
}

@article{schiffman1977role,
  title={The role of number and familiarity of stimuli in the perception of brief temporal intervals},
  author={Schiffman, HR and Bobko, Douglas J},
  journal={The American journal of psychology},
  pages={85--93},
  year={1977},
  publisher={JSTOR}
}

@article{avant1975stimulus,
  title={Stimulus familiarity influences perceived duration in prerecognition visual processing.},
  author={Avant, Lloyd L and Lyman, Paul J},
  journal={Journal of Experimental Psychology: Human Perception and Performance},
  volume={1},
  number={3},
  pages={205--213},
  year={1975},
  publisher={American Psychological Association}
}

@article{jafarpour2017familiarity,
  title={Familiarity expands space and contracts time},
  author={Jafarpour, Anna and Spiers, Hugo},
  journal={Hippocampus},
  volume={27},
  number={1},
  pages={12--16},
  year={2017},
  publisher={Wiley Online Library}
}

@misc{li2026timeblindspatiotemporalcompositionalitybenchmark,
  title={{TimeBlind}: A Spatio-Temporal Compositionality Benchmark for Video {LLMs}}, 
  author={Baiqi Li and Kangyi Zhao and Ce Zhang and Chancharik Mitra and Jean de Dieu Nyandwi and Gedas Bertasius},
  year={2026},
  eprint={2602.00288},
  archivePrefix={arXiv},
  primaryClass={cs.CV},
  url={https://arxiv.org/abs/2602.00288},
}

@inproceedings{li2026othermind,
  title     = {The Other Mind: How Language Models Exhibit Human Temporal Cognition},
  author    = {Li, Lingyu and Yao, Yang and Wang, Yixu and Li, Chunbo and Teng, Yan and Wang, Yingchun},
  booktitle = {Proceedings of the AAAI Conference on Artificial Intelligence},
  volume    = {40},
  pages     = {35598--35607},
  year      = {2026},
  doi       = {10.1609/aaai.v40i42.40871}
}

@inproceedings{de-langis-etal-2025-llms,
    title = "How {LLM}s Comprehend Temporal Meaning in Narratives: A Case Study in Cognitive Evaluation of {LLM}s",
    author = "De Langis, Karin  and
      Park, Jong Inn  and
      Schramm, Andreas  and
      Hu, Bin  and
      Le, Khanh Chi  and
      Kang, Dongyeop",
    editor = "Che, Wanxiang  and
      Nabende, Joyce  and
      Shutova, Ekaterina  and
      Pilehvar, Mohammad Taher",
    booktitle = "Proceedings of the 63rd Annual Meeting of the Association for Computational Linguistics (Volume 1: Long Papers)",
    month = jul,
    year = "2025",
    address = "Vienna, Austria",
    publisher = "Association for Computational Linguistics",
    url = "https://aclanthology.org/2025.acl-long.1415/",
    doi = "10.18653/v1/2025.acl-long.1415",
    pages = "29174--29191",
    ISBN = "979-8-89176-251-0"
}

@inproceedings{liu-etal-2024-tempcompass,
    title = "{T}emp{C}ompass: Do Video {LLM}s Really Understand Videos?",
    author = "Liu, Yuanxin  and
      Li, Shicheng  and
      Liu, Yi  and
      Wang, Yuxiang  and
      Ren, Shuhuai  and
      Li, Lei  and
      Chen, Sishuo  and
      Sun, Xu  and
      Hou, Lu",
    editor = "Ku, Lun-Wei  and
      Martins, Andre  and
      Srikumar, Vivek",
    booktitle = "Findings of the Association for Computational Linguistics: ACL 2024",
    month = aug,
    year = "2024",
    address = "Bangkok, Thailand",
    publisher = "Association for Computational Linguistics",
    url = "https://aclanthology.org/2024.findings-acl.517/",
    doi = "10.18653/v1/2024.findings-acl.517",
    pages = "8731--8772",
}

@techreport{Chatterji2025HowPeopleUse,
  title     = {How People Use {ChatGPT}},
  author    = {Chatterji, Aaron and Cunningham, Thomas and Deming, David J. and Hitzig, Zoë and Ong, Christopher and Shan, Carl Yan and Wadman, Kevin},
  year      = {2025},
  month     = {September},
  type      = {NBER Working Paper},
  number    = {34255},
  institution = {National Bureau of Economic Research},
  doi       = {10.3386/w34255},
  url       = {https://www.nber.org/papers/w34255}
}

@article{skylark2017further,
  title={Further evidence that the effects of repetition on subjective time depend on repetition probability},
  author={Skylark, William J and Gheorghiu, Ana I},
  journal={Frontiers in Psychology},
  volume={8},
  pages={1915},
  year={2017},
  publisher={Frontiers Media SA}
}

@article{birngruber2015introducing,
  title={Introducing a control condition in the classic oddball paradigm: Oddballs are overestimated in duration not only because of their oddness},
  author={Birngruber, Teresa and Schr{\"o}ter, Hannes and Ulrich, Rolf},
  journal={Attention, Perception, \& Psychophysics},
  volume={77},
  number={5},
  pages={1737--1749},
  year={2015},
  publisher={Springer}
}

@article{pariyadath2007effect,
  title={The effect of predictability on subjective duration},
  author={Pariyadath, Vani and Eagleman, David},
  journal={PloS one},
  volume={2},
  number={11},
  pages={e1264},
  year={2007},
  publisher={Public Library of Science San Francisco, USA}
}

@article{eagleman2008human,
  title={Human time perception and its illusions},
  author={Eagleman, David M},
  journal={Current opinion in neurobiology},
  volume={18},
  number={2},
  pages={131--136},
  year={2008},
  publisher={Elsevier}
}

@article{wehrman2020expected,
  title={The expected oddball: Effects of implicit and explicit positional expectation on duration perception},
  author={Wehrman, Jordan J and Wearden, John and Sowman, Paul},
  journal={Psychological research},
  volume={84},
  number={3},
  pages={713--727},
  year={2020},
  publisher={Springer}
}

@inproceedings{zhou2019goingvacationtakeslonger,
    title = "``{G}oing on a vacation'' takes longer than ``{G}oing for a walk'': A Study of Temporal Commonsense Understanding",
    author = "Zhou, Ben  and
      Khashabi, Daniel  and
      Ning, Qiang  and
      Roth, Dan",
    editor = "Inui, Kentaro  and
      Jiang, Jing  and
      Ng, Vincent  and
      Wan, Xiaojun",
    booktitle = "Proceedings of the 2019 Conference on Empirical Methods in Natural Language Processing and the 9th International Joint Conference on Natural Language Processing (EMNLP-IJCNLP)",
    month = nov,
    year = "2019",
    address = "Hong Kong, China",
    publisher = "Association for Computational Linguistics",
    url = "https://aclanthology.org/D19-1332/",
    doi = "10.18653/v1/D19-1332",
    pages = "3363--3369"
}

@inproceedings{chen2021datasetansweringtimesensitivequestions,
 author = {Chen, Wenhu and Wang, Xinyi and Wang, William Yang},
 booktitle = {Proceedings of the Neural Information Processing Systems Track on Datasets and Benchmarks},
 editor = {J. Vanschoren and S. Yeung},
 pages = {},
 title = {A Dataset for Answering Time-Sensitive Questions},
 url = {https://datasets-benchmarks-proceedings.neurips.cc/paper_files/paper/2021/file/1f0e3dad99908345f7439f8ffabdffc4-Paper-round2.pdf},
 volume = {1},
 year = {2021}
}

@inproceedings{tan2023benchmarkingimprovingtemporalreasoning,
    title = "Towards Benchmarking and Improving the Temporal Reasoning Capability of Large Language Models",
    author = "Tan, Qingyu  and
      Ng, Hwee Tou  and
      Bing, Lidong",
    editor = "Rogers, Anna  and
      Boyd-Graber, Jordan  and
      Okazaki, Naoaki",
    booktitle = "Proceedings of the 61st Annual Meeting of the Association for Computational Linguistics (Volume 1: Long Papers)",
    month = jul,
    year = "2023",
    address = "Toronto, Canada",
    publisher = "Association for Computational Linguistics",
    url = "https://aclanthology.org/2023.acl-long.828/",
    doi = "10.18653/v1/2023.acl-long.828",
    pages = "14820--14835"
}

@misc{robyn_speer_2022_7199437,
  author       = {Robyn Speer},
  title        = {rspeer/wordfreq: v3.0},
  month        = sep,
  year         = 2022,
  publisher    = {Zenodo},
  version      = {v3.0.2},
  doi          = {10.5281/zenodo.7199437},
  url          = {https://doi.org/10.5281/zenodo.7199437}
}

@inproceedings{wang2024trambenchmarkingtemporalreasoning,
    title = "{TRAM}: Benchmarking Temporal Reasoning for Large Language Models",
    author = "Wang, Yuqing  and
      Zhao, Yun",
    editor = "Ku, Lun-Wei  and
      Martins, Andre  and
      Srikumar, Vivek",
    booktitle = "Findings of the Association for Computational Linguistics: ACL 2024",
    month = aug,
    year = "2024",
    address = "Bangkok, Thailand",
    publisher = "Association for Computational Linguistics",
    url = "https://aclanthology.org/2024.findings-acl.382/",
    doi = "10.18653/v1/2024.findings-acl.382",
    pages = "6389--6415"
}

@inproceedings{chen-etal-2025-perceive,
    title = "Perceive the Passage of Time: A Systematic Evaluation of Large Language Model in Temporal Relativity",
    author = "Chen, Shuang  and
      Zheng, Yining  and
      Li, Shimin  and
      Cheng, Qinyuan  and
      Qiu, Xipeng",
    editor = "Rambow, Owen  and
      Wanner, Leo  and
      Apidianaki, Marianna  and
      Al-Khalifa, Hend  and
      Eugenio, Barbara Di  and
      Schockaert, Steven",
    booktitle = "Proceedings of the 31st International Conference on Computational Linguistics",
    month = jan,
    year = "2025",
    address = "Abu Dhabi, UAE",
    publisher = "Association for Computational Linguistics",
    url = "https://aclanthology.org/2025.coling-main.554/",
    pages = "8304--8313"
}

@inproceedings{fatemi2024testtimebenchmarkevaluating,
 author = {Fatemi, Bahare and Kazemi, Seyed Mehran and Tsitsulin, Anton and Malkan, Karishma and Yim, Jinyeong and Palowitch, John and Seo, Sungyong and Halcrow, Jonathan and Perozzi, Bryan},
 booktitle = {International Conference on Learning Representations},
 editor = {Y. Yue and A. Garg and N. Peng and F. Sha and R. Yu},
 pages = {94426--94447},
 title = {Test of Time: A Benchmark for Evaluating {LLMs} on Temporal Reasoning},
 url = {https://proceedings.iclr.cc/paper_files/paper/2025/file/eb7295a8bc613b375726659c2ecd6f14-Paper-Conference.pdf},
 volume = {2025},
 year = {2025}
}

@inproceedings{wei-etal-2023-menatqa,
    title = "{M}enat{QA}: A New Dataset for Testing the Temporal Comprehension and Reasoning Abilities of Large Language Models",
    author = "Wei, Yifan  and
      Su, Yisong  and
      Ma, Huanhuan  and
      Yu, Xiaoyan  and
      Lei, Fangyu  and
      Zhang, Yuanzhe  and
      Zhao, Jun  and
      Liu, Kang",
    editor = "Bouamor, Houda  and
      Pino, Juan  and
      Bali, Kalika",
    booktitle = "Findings of the Association for Computational Linguistics: EMNLP 2023",
    month = dec,
    year = "2023",
    address = "Singapore",
    publisher = "Association for Computational Linguistics",
    url = "https://aclanthology.org/2023.findings-emnlp.100/",
    doi = "10.18653/v1/2023.findings-emnlp.100",
    pages = "1434--1447",

}

@inproceedings{su2024timobettertemporalreasoning,
  author={Zhaochen Su and Jun Zhang and Tong Zhu and Xiaoye Qu and Juntao Li and Min Zhang and Yu Cheng},
  title={Timo: Towards Better Temporal Reasoning for Language Models},
  booktitle={First Conference on Language Modeling},
  year={2024},
  url={https://arxiv.org/abs/2406.14192}
}

@article{boltz1998processing,
  title={The processing of temporal and nontemporal information in the remembering of event durations and musical structure.},
  author={Boltz, Marilyn G},
  journal={Journal of experimental psychology: human perception and performance},
  volume={24},
  number={4},
  pages={1087--1104},
  year={1998},
  publisher={American Psychological Association}
}

@misc{qwen3technicalreport,
      title={Qwen3 Technical Report}, 
      author={{Qwen Team}},
      year={2025},
      eprint={2505.09388},
      archivePrefix={arXiv},
      primaryClass={cs.CL},
      url={https://arxiv.org/abs/2505.09388}, 
}

@article{openai_gpt4o_system_card,
  author  = {{OpenAI}},
  title   = {{GPT-4o} System Card},
  year    = {2024},
  journal = {arXiv preprint arXiv:2410.21276},
  url     = {https://arxiv.org/abs/2410.21276}
}

@article{palan2018prolific,
  title={{Prolific.ac}---{A} subject pool for online experiments},
  author={Palan, Stefan and Schitter, Christian},
  journal={Journal of behavioral and experimental finance},
  volume={17},
  pages={22--27},
  year={2018},
  publisher={Elsevier}
}

@misc{gemma_2025,
    title={Gemma 3 Technical Report},
    author={{Gemma Team}},
    year={2025},
    eprint={2503.19786},
    archivePrefix={arXiv},
    primaryClass={cs.CL},
    url={https://arxiv.org/abs/2503.19786}
}

@misc{openai2025gptoss120bgptoss20bmodel,
      title={gpt-oss-120b \& gpt-oss-20b Model Card},
      author={{OpenAI}},
      year={2025},
      eprint={2508.10925},
      archivePrefix={arXiv},
      primaryClass={cs.CL},
      url={https://arxiv.org/abs/2508.10925}, 
}

@misc{openai_gpt54_thinking_system_card,
  title        = {{GPT-5.4 Thinking} System Card},
  author       = {{OpenAI}},
  year         = {2026},
  month        = mar,
  howpublished = {OpenAI Deployment Safety Hub},
  note         = {Includes an appendix on {GPT-5.4 mini}},
  url          = {https://deploymentsafety.openai.com/gpt-5-4-thinking}
}

@misc{openai_gpt55_system_card,
  title        = {{GPT-5.5} System Card},
  author       = {{OpenAI}},
  year         = {2026},
  month        = apr,
  howpublished = {OpenAI Deployment Safety Hub},
  url          = {https://deploymentsafety.openai.com/gpt-5-5}
}

@techreport{anthropic_sonnet_46_system_card,
  title        = {System Card: {Claude Sonnet} 4.6},
  author       = {{Anthropic}},
  institution  = {Anthropic},
  year         = {2026},
  month        = feb,
  url          = {https://www-cdn.anthropic.com/bbd8ef16d70b7a1665f14f306ee88b53f686aa75.pdf}
}

@techreport{anthropic_haiku_45_system_card,
  title        = {{Claude Haiku 4.5} System Card},
  author       = {{Anthropic}},
  institution  = {Anthropic},
  year         = {2025},
  month        = oct,
  url          = {https://www-cdn.anthropic.com/7aad69bf12627d42234e01ee7c36305dc2f6a970.pdf}
}

@misc{grattafiori2024llama3herdmodels,
      title={The {Llama} 3 Herd of Models}, 
      author={Aaron Grattafiori and Abhimanyu Dubey and Abhinav Jauhri and others},
      year={2024},
      eprint={2407.21783},
      archivePrefix={arXiv},
      primaryClass={cs.AI},
      url={https://arxiv.org/abs/2407.21783}, 
}

@inproceedings{pezeshkpour2024large,
  title={Large language models sensitivity to the order of options in multiple-choice questions},
  author={Pezeshkpour, Pouya and Hruschka, Estevam},
  booktitle={Findings of the Association for Computational Linguistics: NAACL 2024},
  pages={2006--2017},
  year={2024}
}

@inproceedings{tam2025none,
    title = "None of the Above, Less of the Right Parallel Patterns in Human and {LLM} Performance on Multi-Choice Questions Answering",
    author = "Tam, Zhi Rui  and
      Wu, Cheng-Kuang  and
      Lin, Chieh-Yen  and
      Chen, Yun-Nung",
    editor = "Che, Wanxiang  and
      Nabende, Joyce  and
      Shutova, Ekaterina  and
      Pilehvar, Mohammad Taher",
    booktitle = "Findings of the Association for Computational Linguistics: ACL 2025",
    month = jul,
    year = "2025",
    address = "Vienna, Austria",
    publisher = "Association for Computational Linguistics",
    url = "https://aclanthology.org/2025.findings-acl.1031/",
    doi = "10.18653/v1/2025.findings-acl.1031",
    pages = "20112--20134",
    ISBN = "979-8-89176-256-5"
}

@inproceedings{li-etal-2020-unqovering,
    title = "{UNQOVER}ing Stereotyping Biases via Underspecified Questions",
    author = "Li, Tao  and
      Khashabi, Daniel  and
      Khot, Tushar  and
      Sabharwal, Ashish  and
      Srikumar, Vivek",
    editor = "Cohn, Trevor  and
      He, Yulan  and
      Liu, Yang",
    booktitle = "Findings of the Association for Computational Linguistics: EMNLP 2020",
    month = nov,
    year = "2020",
    address = "Online",
    publisher = "Association for Computational Linguistics",
    url = "https://aclanthology.org/2020.findings-emnlp.311/",
    doi = "10.18653/v1/2020.findings-emnlp.311",
    pages = "3475--3489"
}

@inproceedings{wolf-etal-2020-transformers,
    title = "Transformers: State-of-the-Art Natural Language Processing",
    author = "Wolf, Thomas  and
      Debut, Lysandre  and
      Sanh, Victor  and
      Chaumond, Julien  and
      Delangue, Clement  and
      Moi, Anthony  and
      Cistac, Pierric  and
      Rault, Tim  and
      Louf, Remi  and
      Funtowicz, Morgan  and
      Davison, Joe  and
      Shleifer, Sam  and
      von Platen, Patrick  and
      Ma, Clara  and
      Jernite, Yacine  and
      Plu, Julien  and
      Xu, Canwen  and
      Le Scao, Teven  and
      Gugger, Sylvain  and
      Drame, Mariama  and
      Lhoest, Quentin  and
      Rush, Alexander",
    editor = "Liu, Qun  and
      Schlangen, David",
    booktitle = "Proceedings of the 2020 Conference on Empirical Methods in Natural Language Processing: System Demonstrations",
    month = oct,
    year = "2020",
    address = "Online",
    publisher = "Association for Computational Linguistics",
    url = "https://aclanthology.org/2020.emnlp-demos.6/",
    doi = "10.18653/v1/2020.emnlp-demos.6",
    pages = "38--45"
}

@inproceedings{paszke2019pytorch,
  title     = {{PyTorch}: An Imperative Style, High-Performance Deep Learning Library},
  author    = {Paszke, Adam and Gross, Sam and Massa, Francisco and Lerer, Adam and Bradbury, James and Chanan, Gregory and Killeen, Trevor and Lin, Zeming and Gimelshein, Natalia and Antiga, Luca and Desmaison, Alban and K{\"o}pf, Andreas and Yang, Edward and DeVito, Zachary and Raison, Martin and Tejani, Alykhan and Chilamkurthy, Sasank and Steiner, Benoit and Fang, Lu and Bai, Junjie and Chintala, Soumith},
  booktitle = {Advances in Neural Information Processing Systems},
  volume    = {32},
  pages     = {8024--8035},
  year      = {2019}
}

@inproceedings{dettmers2023qlora,
  title     = {{QLoRA}: Efficient Finetuning of Quantized {LLMs}},
  author    = {Dettmers, Tim and Pagnoni, Artidoro and Holtzman, Ari and Zettlemoyer, Luke},
  booktitle = {Advances in Neural Information Processing Systems},
  volume    = {36},
  pages     = {10088--10115},
  year      = {2023}
}

@techreport{radford2019language,
  title       = {Language Models are Unsupervised Multitask Learners},
  author      = {Radford, Alec and Wu, Jeffrey and Child, Rewon and Luan, David and Amodei, Dario and Sutskever, Ilya},
  institution = {OpenAI},
  year        = {2019},
  url         = {https://cdn.openai.com/better-language-models/language_models_are_unsupervised_multitask_learners.pdf}
}

@misc{honnibal2020spacy,
  title     = {{spaCy}: Industrial-strength Natural Language Processing in {P}ython},
  author    = {Honnibal, Matthew and Montani, Ines and Van Landeghem, Sofie and Boyd, Adriane},
  year      = {2020},
  publisher = {Zenodo},
  doi       = {10.5281/zenodo.1212303},
  url       = {https://doi.org/10.5281/zenodo.1212303}
}

@book{bird2009natural,
  title     = {Natural Language Processing with {P}ython: Analyzing Text with the Natural Language Toolkit},
  author    = {Bird, Steven and Klein, Ewan and Loper, Edward},
  publisher = {O'Reilly Media},
  address   = {Sebastopol, CA},
  year      = {2009}
}

\appendix
\raggedbottom
\makeatletter\if@twocolumn\fi\makeatother
\onecolumn
\newpage 
\twocolumn
\section{Additional Benchmark Information}
\label{appendix:Appendix_dataset}

\subsection{Full Benchmark Design}
\label{sec:benchmark_design_full}

By translating classic psychology experiments into narrative prompts, we’re able to mimic how the human brain stretches and compresses intervals through writing. This lets us easily evaluate how LLMs mirror or diverge from human-like temporal biases. Each of the five illusions has multiple question styles mimicking different types of evaluations and components in classic psychology. Most templates contain both longer and shorter versions of the question to account for the effects of passage length on perceived time if allowed by the question design. We outline a detailed explanation of each template type and its reasoning below; for concrete examples, refer to Table~\ref{tab:template_examples}.

\subsubsection{Emotional Time Dilation}

Emotional time dilation, which occurs with high-arousal stimuli, causes an individual to perceive an interval as longer than the truth. For example, people with a fear of spiders overestimate the duration of a brief period spent watching a spider in comparison to people without that fear~\cite{watts1984fear}. Other studies targeting this illusion used mutilated images to stimulate arousal. The mutilated images showed significant overestimation compared to neutral faces~\cite{gil2012emotional}. The pleasantness of an event and arousal influence both real-time perception and retrospective memory encoding~\cite{hurlemann2005noradrenergic}.

We translated the experiments into a narrative framework that uses two variables: situational stakes and narrative explication. We create conditions that isolate the factors. While the literature generally associates emotional time dilation with stimulus-driven surprise or threat, we adopt a broader definition that considers an arousal–attention system. Our conditions also examine context-induced arousal (via stakes) and attentional load as emotional mechanisms known to influence perceived duration.

We distinguish this from the Oddball Effect by the source of attention (top-down vs.\ bottom-up; see \S~\ref{sec:design_principles}).

\begin{itemize}
    \item[A -] Stakes and Expectations: The same neutral stimulus (e.g., a flickering reflection) appears during either a low-stakes or a high-stakes activity. Because only the situational stakes change, this condition tests whether potential consequences or pressure alter how the character engages with an identical stimulus, and in turn how long it feels.
    
    \item[B -] Interval Intensity: While waiting through an otherwise empty interval, the character does either a high-focus task or a low-demand activity. The high-focus task induces cognitive arousal: when cognitive resources are heavily engaged, fewer remain available for tracking time, altering how long the interval feels~\cite{brown2014time}.
    
    \item [C -] Emotional Arousal: With context and stakes held constant, the character witnesses either a neutral baseline event or a sudden, unexpected, or emotional one. This tests whether involuntary attentional capture from surprise or emotional salience expands perceived duration.
\end{itemize}
	
By introducing high-stakes or threatening elements, we simulate the arousal in emotional time dilation studies through narrative events. By holding the objective action constant while varying the emotional context, we can test whether participants perceive the same event as lasting longer when the scene feels more intense.

In laboratory settings, researchers simulate emotional time dilation in the participant directly by inducing an increase in heart rate. In contrast, our narrative-based questions ask participants to imagine these states rather than experience them directly. As a result, responses may reflect reasoning about what should feel longer, rather than reproducing how an internal timing mechanism would actually behave under real arousal.

\subsubsection{Oddball Effect}

The Oddball Effect happens when a novel stimulus appears within a series of identical stimuli. This makes the unique item seem to last longer~\cite{pariyadath2007effect}. Duration overestimation is more pronounced for the oddball~\cite{birngruber2015introducing}. Two contrasting mechanisms drive this illusion: a top-down process, where the closer an individual gets to expecting a change, the more significant that change feels when it occurs; and a bottom-up process, where the rarity of the oddball automatically captures attention and interrupts the brain's habituation to the repeated sequence~\cite{wehrman2020expected}. We mimicked the effects of the experiments by using sequential predictability and novel stimuli. 

\begin{itemize}
    \item[A -] Sequence Disruption: We create this illusion with a predictable sequence, and then introduce a deviation within that pattern. The disruption captures attention through bottom-up novelty, mirroring classic oddball paradigms where a distinct stimulus within a uniform sequence is perceived as lasting longer. We divide this illusion template into three subsections:
    \begin{itemize}
        \item[1.] Containing: The oddball occurs within the repeating sequence itself, embedded directly in the interval being evaluated.
        \item[2.] Preceding: The oddball appears before the target interval and influences how the following interval is perceived.
        \item[3.] Following: The oddball occurs after the target interval, testing how a later disruption retroactively affects the perceived duration.
    \end{itemize}
\end{itemize}

To separate the overlap between the Oddball Effect and Emotional Time Dilation, we distinguish structural surprise and arousal. The Oddball Effect relies on perceptual novelty within a sequence. It does not require the ``oddball'' to be scary, exciting, or high-stakes; it simply needs to be different from the preceding standards. 

\subsubsection{Filled-Duration Effect}

The Filled-Duration Effect describes when a time interval containing discrete elements is perceived as longer than an empty interval of an objectively identical duration. In experimental settings, designs consistently show that intervals containing brief tones or visual markers appear longer than silent gaps~\cite{thomas1974filled}. The number of elements affects this. The more distinct events the brain processes within the window, the longer the window of time feels~\cite{buffardi1971factors}. In parallel, continuous structured or “melodic” sequences are shorter than silent ones, showing that time flies when listening to music tested over longer durations~\cite{droit2010time}. 

To translate these findings into a narrative framework, we fill an interval by comparing a sequence of distinct actions to a continuous period of time:

\begin{itemize}
    \item [A -] Information Density: We contrast an interval filled with events with an empty waiting period, reflecting findings that filled intervals are often perceived as longer than empty ones for short durations.
\end{itemize}

From the reader's perspective, we expect the filled interval to feel longer. However, from the character's perspective, the empty interval may feel longer, as the absence of stimulation can heighten awareness of time passing (similar to how engaging activities, like music, can make longer durations feel shorter). This structure mirrors existing literature on short- versus long-duration judgments by asking participants to evaluate duration from both an external and internal perspective.

\subsubsection{Familiarity Duration Effect}
The subjective duration of an interval depends on whether the brain recognizes the task. This phenomenon is known as the familiarity-duration effect.  Generally, novel events are perceived as lasting longer than familiar ones, where the newness of an experience needs more cognitive processing and more frequent neural sampling~\cite{manahova2020familiarity, avant1975stimulus}. When a task becomes routine, the brain processes certain recurring activities more effectively, and time will begin to pass more quickly~\cite{skylark2017further}.

We focus on the cognitive differences between novelty and routine to demonstrate this illusion. We define familiarity in two ways: 1) the general perceived familiarity by the public, and 2) the declared expertise of the character.

\begin{itemize}
    \item[A - ] Labeled Expertise: We present identical steps but explicitly frame the event as either “familiar” or “unfamiliar.” This tests whether labeling alone influences perceived duration, targeting whether participants infer temporal perception without changing the core actions.
    \item[B - ] General Novelty: We compare a character performing an objectively familiar, everyday task with an objectively unfamiliar task, while holding the character’s stated level of familiarity constant across both. This tests whether intrinsic task novelty affects participants' perceived duration.
\end{itemize}

A significant challenge and possible shortcoming in translating familiarity effects into a test question set is that ``familiarity'' is not universal. It can be culturally and demographically dependent. A laboratory study can ensure novelty by using controlled stimuli; narrative prompts rely on real-world concepts. What is a routine, ``compressed'' task for a digital native might be a high-effort, ``expanded'' task for an older adult or someone from a different professional background. 

\subsubsection{Temporal Order Effects}

Temporal Order Effects describe how the structure and predictability of event order influence subjective duration. We distinguish between two complementary processes: order violation and order redundancy. A person needs to actively pay attention to the sequence to detect an explicit order violation. This can cause duration judgments to become overestimated~\cite{brown2014time}. Actively anticipating or monitoring for an upcoming event can lengthen perceived duration~\cite{skylark2017further}. Order redundancy happens when repeated stimuli compress subjective time. To translate these sequence effects, we isolate the order of a character’s workflow:

\begin{itemize}
    \item[A -] Scrambled Workflow: We present a character performing a task in the wrong order, explicitly stating it at the start. This forces participants to be aware of the incorrect sequence throughout. This mirrors experimental setups showing that violations of expected order reduce temporal accuracy and alter perceived duration. This structure follows~\citet{skylark2017further}.
    \item[B -] Event Repetition: We compare a narrative sequence containing repetitive actions to a sequence with distinct events. This tests whether repetition within a sequence compresses perceived duration, as repeated events become more predictable and require less processing, leading to a denser, more compressed memory representation of the interval. This manipulation does not introduce an outlier. It reduces novelty across the sequence with repetition. 
    \item[C -] Narrative Disruption: We present a sequential, coherent narrative that an unexpected event suddenly interrupts. This tests whether an unexpected event triggers retrospective expansion of the surrounding interval, as participants re-evaluate prior events in light of the disruption. The effect comes from a structural violation of the sequence, leading to reinterpretation of what came before, rather than from physiological or emotional arousal. 
\end{itemize}

\subsection{Dataset Creation}
Each template consists of a paired control and experimental condition that differs in a targeted manipulation. The experimental condition corresponds to the scenario that, based on prior psychological literature, is expected to be perceived as longer for the character. We design the templates so that the content closely matches in length, wording, and narrative structure.

We design our collection of fillers (stimuli) for contextual variation while preserving the structure of each illusion type. We wrote the initial set of fillers per category manually and then expanded it using GPT-4o~\cite{openai_gpt4o_system_card}. We subsequently reviewed and edited all generated fillers by hand. List sizes vary with what each slot must hold. Open slots, such as events, empty intervals, and predictable sequences, have 20--31 entries. Slots that must stay matched across conditions are kept small so each pair stays closely matched: the paired low- and high-stakes activities and the high-focus and neutral tasks in Emotional Time Dilation have 5 entries each. The time-context phrases in Filled Duration (e.g., ``the evening'', ``the commute'') have 3--8 entries per setting, since few natural phrases avoid explicit duration markers. Multi-step tasks are stored as index-aligned lists, so each step belongs to its task: 10 tasks per context, familiarity, and length cell for Familiarity, and 100 workflows per length for Temporal Order. We sampled 50 character names at random from the top 100 most common names in the United States over the last one hundred years~\cite{ssa_babynames_century}. Using these fillers, we generated 2000 paired scenarios for each illusion type by randomly sampling from the available templates, resulting in a total of 10000 candidate paired scenarios. In addition to the filtered dataset, we manually curated a gold set of 250 items (50 per illusion type). We selected and lightly edited these items by hand to ensure clarity, consistency, and faithful representation of each illusion.

To reduce stylistic noise and increase grammatical coherence, we paraphrased each paired scenario three times using Qwen3-32B~\cite{qwen3technicalreport} with gold items as few-shot examples (Appendix~\ref{appendix:Appendix_dataset}). From these variants, we selected the version that most closely matched the desired length constraints between control and experimental conditions for evaluation. We designed this step to minimize differences in word count that could bias perceived duration judgments. To filter out paraphrases that drifted from their source, we compared the word count of each paraphrased passage with that of the same passage before paraphrasing. We kept a combination if at least one of its paraphrased passages was within seven words of its source and removed it if every paraphrased passage differed by more than seven words. This rule bounds drift in length only; it does not bound the gap between a pair's experimental and control passages, which the variant selection above minimizes, and the selected variant is not itself required to fall within seven words of its source. After filtering, 6,684 paired scenarios remained, with 3316 removed across all categories (Table~\ref{tab:dataset_sizes}).

Because we generate items from a small number of templates, examples within a template family are not statistically independent; accounting for this, the effective sample size is smaller.

\begin{table}[ht]
\centering
\small
\setlength{\tabcolsep}{4pt}
\renewcommand{\arraystretch}{1.2}
\begin{tabular}{lrrr}
\toprule
\textbf{Illusion Type} & \textbf{Before} & \textbf{After} & \textbf{Removed} \\
\midrule
emotional time dilation & 2000 & 1630 & 370 \\
oddball effect & 2000 & 1208 & 792 \\
filled duration effect & 2000 & 1563 & 437 \\
familiarity duration effect & 2000 & 1212 & 788 \\
temporal order & 2000 & 1071 & 929 \\
\midrule
\textbf{Total} & \textbf{10000} & \textbf{6684} & \textbf{3316} \\
\bottomrule
\end{tabular}
\caption{Dataset size before and after filtering.}
\label{tab:dataset_sizes}
\end{table}

While the dataset is not perfectly balanced after filtering, each category retains a sufficiently large sample size for analysis (Table~\ref{tab:subtype_breakdown}). Although this imbalance may introduce minor differences in statistical power across conditions, it does not pose a significant limitation for our evaluation. Future work may re-balance categories. 

We will make the complete generated dataset, slot fillers, anonymized human responses, and scripts available in a public repository under the MIT License.\footnote{\url{https://github.com/utahnlp/temporal-illusions}}
The released data includes model-generated text (fillers expanded with GPT-4o through the OpenAI API) and the terms give ownership of outputs to the user, and paraphrases generated by Qwen3-32B, which is released under the Apache 2.0 license. Character names come from public data published by the U.S. Social Security Administration. Because arousal is the manipulation, some Emotional Time Dilation items describe mildly arousing or threatening events by design (e.g., a clown chasing a child), but no item contains offensive, hateful, or explicit content. 

\begin{table}[htbp]
\centering
\small
\setlength{\tabcolsep}{4pt}
\renewcommand{\arraystretch}{1.0}
\begin{tabular}{lrr}
\toprule
\textbf{Illusion Type/Subtype} & \textbf{Count} & \textbf{\%} \\
\midrule
\multicolumn{3}{c}{\em Emotional Time Dilation}\\
Interval Intensity & 721 & 44.2\% \\
Stakes and Expectations & 629 & 38.6\% \\
Emotional Arousal & 280 & 17.2\% \\
\midrule

\multicolumn{3}{c}{\em Oddball Effect}\\
Containing & 427 & 35.3\% \\
Following & 398 & 32.9\% \\
Preceding & 383 & 31.7\% \\
\midrule

\multicolumn{3}{c}{\em Filled Duration Effect}\\
Information Density & 1563 & 100.0\% \\
\midrule

\multicolumn{3}{c}{\em Familiarity Duration Effect}\\
General Novelty & 619 & 51.1\% \\
Labeled Expertise & 593 & 48.9\% \\
\midrule

\multicolumn{3}{c}{\em Temporal Order Effect}\\
Scrambled Workflow & 489 & 45.7\% \\
Narrative Disruption & 355 & 33.1\% \\
Event Repetition & 227 & 21.2\% \\
\bottomrule

\end{tabular}

\caption{Distribution of paired scenarios across illusion types and subtypes after filtering.}
\label{tab:subtype_breakdown}
\end{table}

\begin{table*}[p]
\centering
\small
\renewcommand{\arraystretch}{1.3}
\caption{Example of template structures for each illusion type and corresponding control and experimental conditions}
\label{tab:template_examples}
\begin{tabular}{|p{2.6cm}|p{6cm}|p{6cm}|}
\hline
\textbf{Template} & \textbf{Control} & \textbf{Experimental} \\
\hline

\multicolumn{3}{|c|}{\textbf{Emotional Time Dilation}} \\
\hline
Stakes and Expectations
& Andrew was \textbf{casually checking his email} when he noticed a flickering reflection in the puddle outside. He continued calmly.
& Andrew was \textbf{working on a time-sensitive}, crucial project when he saw a flickering reflection in the puddle outside. He kept working. \\
\hline

Interval Intensity
& While waiting for a delayed train, \textbf{Sarah looked over some unread text messages}.
& While waiting for a delayed train, \textbf{Sarah tried to complete a difficult assignment}. \\
\hline

Emotional Arousal
& Sitting on the bleachers watching a game alone, Richard observed a \textbf{dry leaf rolling along the pavement, scraped by a light breeze that spun it and paused before it tumbled again toward the gutter}. He continued with his own business.
& Sitting on the bleachers watching a game alone, Richard noticed a \textbf{man wearing bright post-it notes walking down the sidewalk. A few post-its flew in the wind before he vanished around the corner}. He continued with his own business. \\
\hline

\multicolumn{3}{|c|}{\textbf{Oddball Effect}} \\
\hline
Sequence Disruption
& James systematically processed badge activation processes, ensuring each setup matched perfectly. \textbf{Nothing was different.} He finished all tasks.
& James systematically processed badge activation processes, but a \textbf{peculiar channel pattern emerged midway}. He continued and finished all tasks. \\
\hline

\multicolumn{3}{|c|}{\textbf{Filled-Duration Effect}} \\
\hline
Information Density
& Emily spent the evening at home with various tasks. Throughout it, \textbf{she polished silverware, ironed tablecloths, arranged centerpieces, and set place settings}. She grabbed a snack after. 
& Emily spent the evening at home with little to do. Throughout it, \textbf{she lay on the floor watching the ceiling fan spin}. Eventually, she was hungry enough to grab a snack.\\
\hline

\multicolumn{3}{|c|}{\textbf{Familiarity Duration Effect}} \\
\hline
Labeled Expertise
& David started building a detailed model airplane during his leisure time, naturally \textbf{following familiar steps in order}: first sorting pieces, then assembling the fuselage and wings, next adding small components, and finally inspecting the model.
& David began constructing a detailed model airplane during his free time, carefully \textbf{following each unfamiliar step}: first sorting pieces, then assembling the fuselage and wings, next adding small components, and finally inspecting the model. \\
\hline

General Novelty
& At home, Linda began \textbf{boiling water} using these steps: filling a kettle with water, placing it on the stove, heating until it boils, then turning off once boiling is confirmed.
& Linda \textbf{assembled a complex bookshelf} for \textbf{the first time}: unpacking components, arranging parts to identify screws and connectors, checking stability, and making adjustments. \\
\hline

\multicolumn{3}{|c|}{\textbf{Temporal Order Effects}} \\
\hline
Scrambled Workflow
& Melissa tied her shoelaces methodically and efficiently. First, she grabbed the two laces. Next, crossing and pulling them tight. Then, form a loop and thread the other lace. Finally, pulling the knot tight and securing it. Finishing each step.
& Melissa was tying her shoelaces but realized \textbf{she was doing it incorrectly. First, Melissa crossed and pulled the laces tight.} Next, she grabbed them. Then, pulling the knot tight and securing it. Finally, she formed a loop and threaded the other lace. \\
\hline

Event Repetition
& Joshua made a list of recent tasks: running, \textbf{bathroom break}, resting after work, and \textbf{a second bathroom visit}.
&  Joshua noted down activities like going to the bathroom, running, and relaxing after a long day by watching a basketball game.\\
\hline

Narrative Disruption
& Starting the day, Richard makes his bed. Then, he reads a book. Afterwards, he goes on a run. Finally, he continues to read the book he read earlier.
& Starting the day, Richard makes his bed. Next, he reads a book. Suddenly, \textbf{he sees a man covered in Post-it notes walking silently.} Finally, he goes for a run. \\
\hline

\end{tabular}
\end{table*}

\begin{figure*}[t]
\begin{tcolorbox}[title=Full Paraphrasing Prompt, colback=white, colframe=black, fontupper=\small]
\label{sec:paraphrasing_prompt}

\textit{We insert the examples after the system prompt and before the final user query, as alternating USER/ASSISTANT turns (i.e., standard few-shot formatting).}

\begin{verbatim}
SYSTEM:
You are an expert editor specializing in paraphrasing sentence pairs for psychology research.
PRIORITIES (strictly in this order):
1. The experimental and control fields must each contain a full, fluent sentence. Never use 
placeholder words like experimental or control as the sentence text.
2. Grammatical correctness and sentence fluency — output must sound completely natural
3. Matched lengths — experimental and control sentences must be as close in word count as 
possible
4. Core meaning — preserve the key semantic contrast between sentences; details may be omitted 
when needed for fluency or length
Return ONLY a valid JSON array of exactly 3 paraphrased variants. No preamble, no explanation.

USER:
Paraphrase these sentences. Target ~{N} words each.
Experimental ({ew} words): "{exp sentence}"
Control ({cw} words): "{ctrl sentence}"
Return JSON array of 3 diverse variants:

ASSISTANT:
[{"experimental": "...", "control": "..."}, ...]

[... one USER/ASSISTANT pair per few-shot example (2 by default), each drawn from the gold 
set ...]

USER:
Paraphrase these sentences. Target ~{N} words each.
Experimental ({ew} words): "{exp sentence}"
Control ({cw} words): "{ctrl sentence}"
Return JSON array of 3 diverse variants:

ASSISTANT:
\end{verbatim}
\end{tcolorbox}
\end{figure*}

\section{Additional Evaluation Information}
\label{appendix:Appendix_eval}
\subsection{Evaluation Instructions and Prompts}
\label{sec:eval_instructions}

\begin{tcolorbox}[breakable, colback=white, colframe=black, sharp corners, boxrule=0.4pt, boxsep=0pt, left=4pt, right=4pt, top=4pt, bottom=4pt, before upper={\parindent=0pt}, fontupper=\small]
\textbf{Survey Instructions}
\label{sec:instructions}

In this survey, our goal is to understand how people perceive and experience the passage of time. You will be presented with 25 pairs of scenarios, each describing a character undergoing a particular event. For each pair, you will be asked to answer a question comparing the two scenarios.

We are interested in your intuitive, first-instinct responses. Please answer each question as quickly as possible, ideally within 15 seconds. There are no correct or incorrect answers.

Each pair of scenarios describes events that lasted the same objective amount of time. However, your task is to judge how long they \textit{felt}, depending on the perspective specified in the question.

\textbf{Character's Perspective.} \\
\textit{“Given the two events below that lasted the same amount of time, which one do you think felt longer for the character experiencing it?”} \\
Imagine yourself as the character. Which event would feel longer to experience within the story?

\textbf{Reader's Perspective.} \\
\textit{“Given the two events below that lasted the same amount of time, which one felt longer to you while reading?”} \\
Answer from your perspective as the reader. Which event felt longer or more time-consuming to read and process?

Please base your answers on your immediate impression rather than careful deliberation.
\end{tcolorbox}

\begin{tcolorbox}[title=Evaluation Prompts, colback=white, colframe=black, breakable, fontupper=\small]
\label{sec:evaluation_prompts}

\begin{minipage}{\linewidth}
\textbf{Character's Perspective — System}
\begin{verbatim}
You are reading two short stories. 
Imagine you are the character. 
Which event felt longer to
experience within the story?

Given the two following events 
below that lasted the same amount 
of time, which one do you think 
felt longer for the character 
in the events?
\end{verbatim}
\end{minipage}\par\medskip

\begin{minipage}{\linewidth}
\textbf{Reader's Perspective — System}
\begin{verbatim}
You are reading two short stories. 
As the person reading the text, 
which event felt longer or more 
time-consuming to read and process?

Given the two following events below 
that lasted the same amount of time,
which one felt longer to you while 
reading?
\end{verbatim}
\end{minipage}\par\medskip

\begin{minipage}{\linewidth}
\textbf{User Prompt (shared across perspectives)}
\begin{verbatim}
Sentence One: "{sentence}"
Sentence Two: "{sentence}"

Please respond with only "Sentence 
One" or "Sentence Two".
\end{verbatim}
\end{minipage}\par\medskip

\textit{The model selects one passage. We map responses ("Sentence One"/"Sentence Two") back to experimental or control conditions based on presentation order. We randomize passage order once per pair and hold it constant across both perspectives; each pair in the full dataset is run in one order only (gold items in both orders), and positional bias is corrected by symmetrizing across items with the experimental passage first vs.\ second.}

\end{tcolorbox}

\subsection{Configuration Details}
\label{sec:config}
We ran open-weight models locally with HuggingFace Transformers 5.8.0~\cite{wolf-etal-2020-transformers} and PyTorch 2.14.0~\cite{paszke2019pytorch} on NVIDIA L40S GPUs, loaded in 4-bit FP4 precision via bitsandbytes~\cite{dettmers2023qlora} with bfloat16 compute (gpt-oss models retain their native MXFP4 quantization). We used greedy decoding and a 512-token generation budget, using each model's own chat template, with thinking enabled for Qwen3 and gpt-oss (Gemma 3 and Llama 3 have no thinking mode). We ran proprietary models through the OpenAI Batch API and Anthropic Message Batches at default sampling settings and a 1024-token completion budget, with thinking off for Claude and each GPT model's API default reasoning effort (\emph{none} for GPT-5.4 mini and GPT-5.4, \emph{medium} for GPT-5.5). We randomize scenario order per item under a fixed seed (42); each item in the full dataset was run in one order only, and each gold item in both orders. To check the one-order design, we kept one random order per gold item and recomputed $p_{\text{debiased}}$ (200 draws): the estimate was unbiased (mean difference from the both-order estimate at most 0.008 for any model and condition) and the cross-model means varied with a standard deviation of at most 0.017. If an open-weight model's output did not name exactly one option, we re-prompted it with its previous output and a request to answer with only one option, up to nine times (ten attempts in total). Proprietary models received a single attempt with no re-prompting. On the gold set, 1.3\% (Qwen3-8B) and 3.2\% (Qwen3-32B) of saved outputs reached the 512-token budget before the thinking block closed, and 22\% and 10\% of their trials were retried, mostly after a truncated first attempt. gpt-oss outputs stayed under 300 tokens. Re-parsing only the final answers and dropping retried or truncated Qwen3 trials changes the gold-set cross-model means by at most 0.02. GPU computation hours were not tracked.

\subsection{Statistical Corrections}
\label{sec:stats}
Since we instantiate all 6,684 paired scenarios in the data from 115 templates, examples in a template family are not statistically independent. Treating them as such could understate uncertainty. For the LLM evaluation, we cluster standard errors in the originating template, with a Student-t small-sample adjustment for illusions with fewer template varieties. For the human study, we fit a mixed-effects logistic regression with crossed random intercepts for participant and template, which accounts for repeated judgments from the same participant and for items generated from the same template.

\paragraph{Human--LLM comparison.} We used gold items and fit a mixed-effects logistic regression predicting selection of the experimental scenario. Fixed effects included an intercept for each illusion--perspective condition (human log-odds), condition~$\times$~LLM interactions (model--human differences), and group-specific presentation-order effects. The random effects were:
(1) an intercept and an order slope for each rater (participant or model), with separate variances for humans and models;
(2) an intercept for each template, estimated separately for humans and for models, with the two allowed to correlate;
and (3) an intercept for each model--template pair, because each model judges several items from the same template.
We measured every model--human gap in the direction the literature predicts. For filled duration from the reader's perspective, the literature predicts the control (filled) interval, so there the gap is measured toward the control. We tested each condition with likelihood-ratio tests and Benjamini--Hochberg correction across 10 conditions, overall agreement with a 10-df likelihood-ratio test ($\chi^2(10)=130.2$, $p<.001$), and the mean difference with a Wald test. 

\subsection{Evaluation Framework and Dataset Validation}

\begin{figure}[t]
    \centering

    \includegraphics[width=\columnwidth]{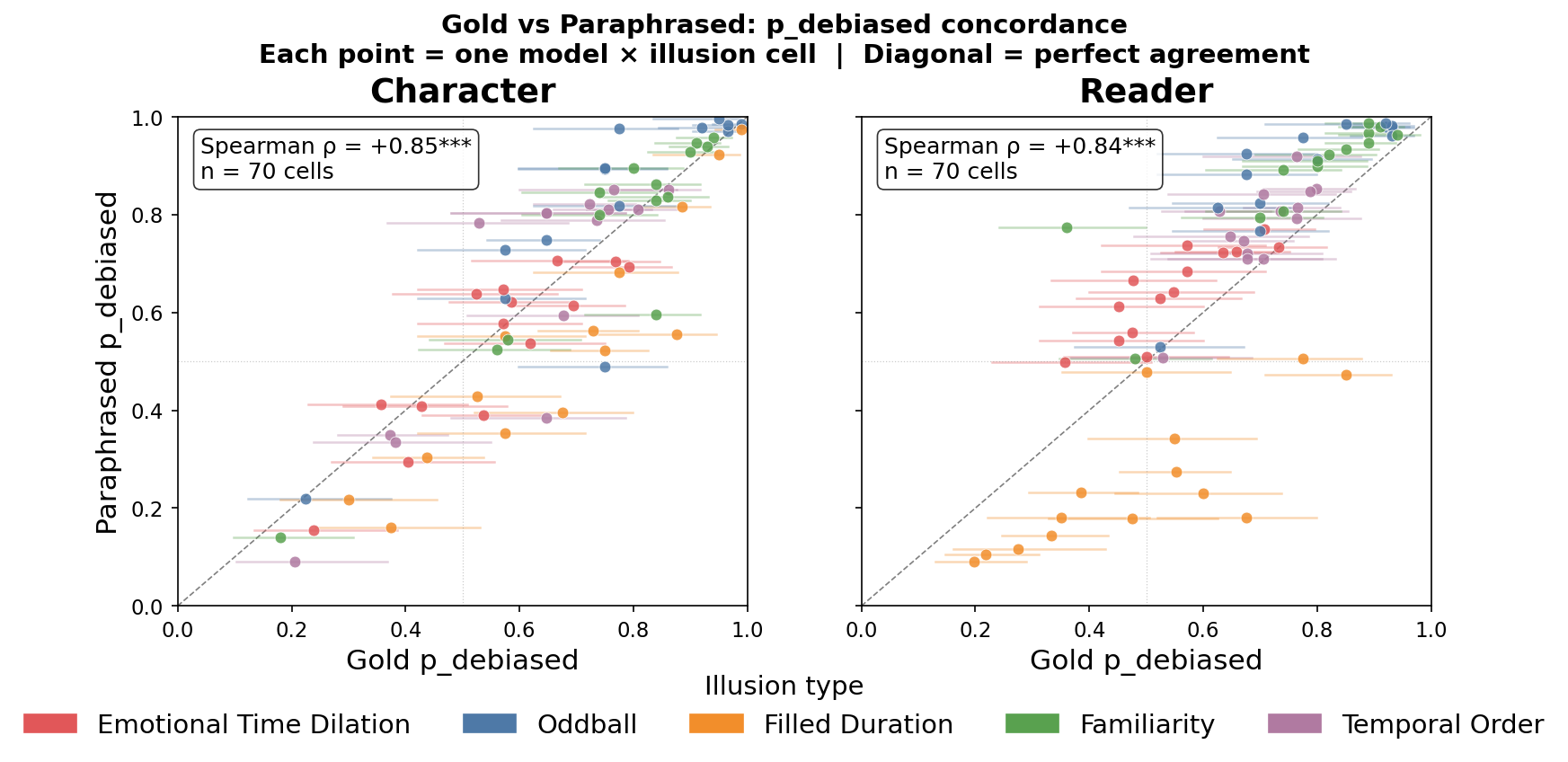}
    \caption*{(a) 4-way evaluation}

    \includegraphics[width=\columnwidth]{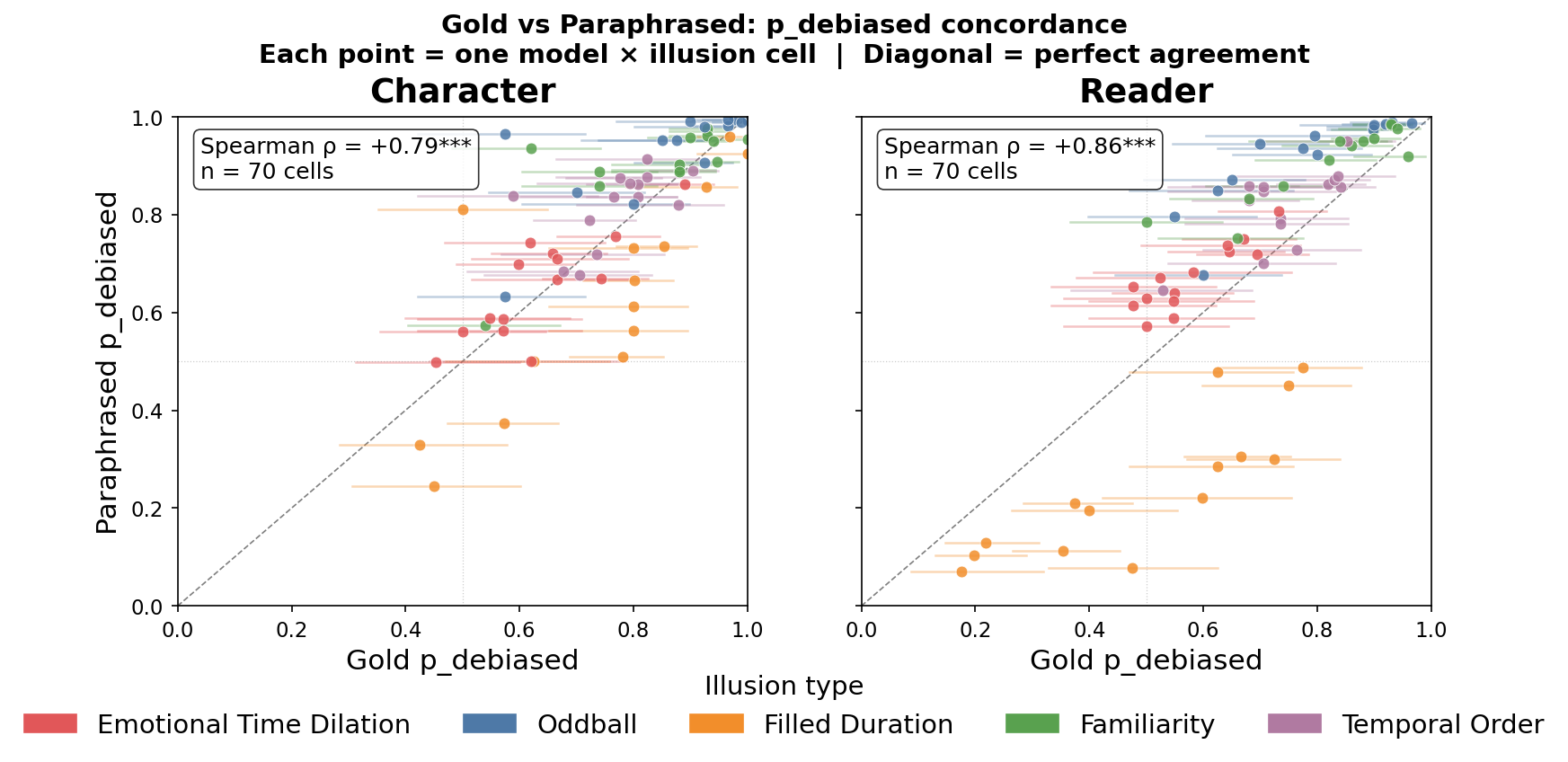}
    \caption*{(b) 2-way evaluation}

    \caption{
    Concordance between the gold set and the paraphrased dataset measured by the Pearson correlation coefficient $r$ and the Spearman rank correlation $\rho$. Each point represents a model–illusion pair; the diagonal indicates perfect agreement.
    }
    \label{fig:concordance}
\end{figure}

\begin{figure}[t]
    \centering

    \includegraphics[width=\columnwidth]{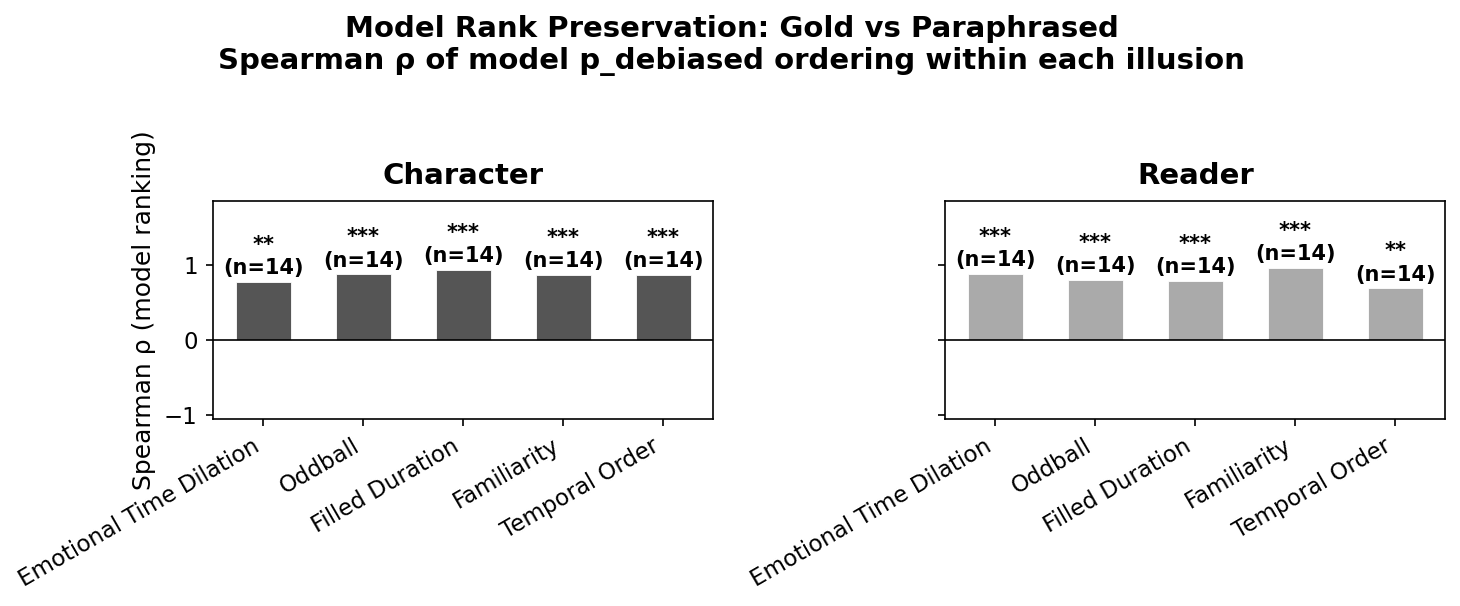}
    \caption*{(a) 4-way evaluation}

    \includegraphics[width=\columnwidth]{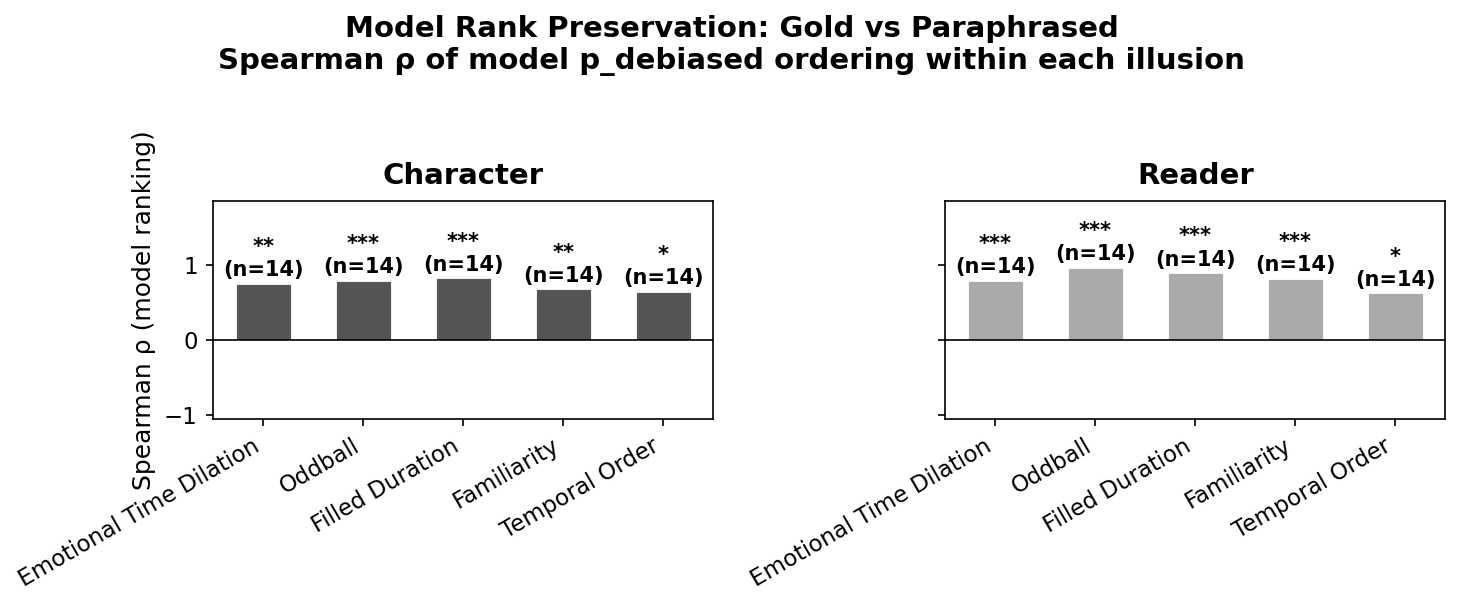}
    \caption*{(b) 2-way evaluation}

    \caption{
    Model rank preservation between the gold set and the paraphrased dataset measured by the Spearman rank correlation $\rho$ across illusion types.    }
    \label{fig:rank}
\end{figure}
The gold set contains 50 hand-edited items per illusion, rewritten to most accurately reflect the illusion design, and designed from the character's perspective. We also create a large automated dataset based on the gold set to enable scalable evaluation.

To assess the consistency between these datasets, we compare model and illusion-level results using two complementary measures: the Pearson correlation coefficient $r$ to assess the linear agreement in effect sizes (i.e., whether the debiased proportions $p_\text{debiased}$ track proportionally across datasets (Figure~\ref{fig:concordance})), and the Spearman rank correlation $\rho$ both to confirm this under rank-based assumptions and to evaluate model rank preservation within each illusion (Figure~\ref{fig:rank}). Overall, the paraphrased dataset provides a strong approximation of the gold set for both perspectives. Illusion-level effect sizes and model rankings are highly consistent for both evaluations (2-way and 4-way).
The 4-way evaluation is also more sensitive to illusion-specific differences than the 2-way evaluation.

Agreement is similar for the reader's perspective (Spearman $\rho = 0.86/0.84$ for 2-way/4-way) and the character's perspective ($\rho = 0.79/0.85$).

Comparing the two evaluations, the 4-way evaluation demonstrates higher fidelity to the gold set for the character's perspective and greater sensitivity to illusion-specific differences. By contrast, the 2-way evaluation provides a simpler approach but reduces sensitivity in more challenging settings. Taken together, these results suggest that 2-way is sufficient for coarse comparisons, whereas 4-way evaluation remains preferable for fine-grained analysis and for capturing subtle behavioral differences across models and illusion types. We perform all analyses for both the 4-way and the 2-way evaluation.

\subsection{Effects of Additional Factors: Positional Bias, Complexity, and Non-Committal Answers}
\label{sec:additional-results}

\begin{figure*}[!htbp]
    \centering
    \includegraphics[width=0.85\textwidth]{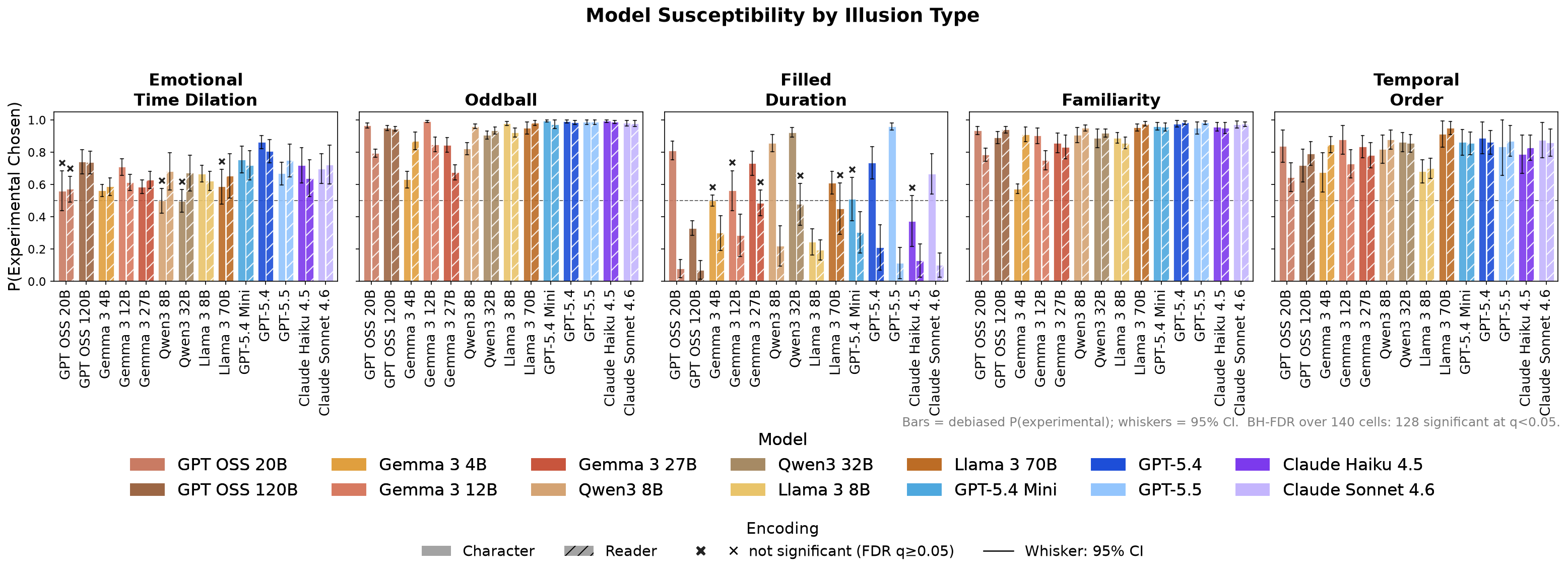}
    
    \caption{
    Model performance across illusion types for 2-way evaluation with positional-bias correction. Bars represent the debiased proportion $p_{\text{debiased}}$ for each model and perspective (character's vs.\ reader's); whiskers are 95\% cluster-robust confidence intervals. Cells marked $\times$ are not significant under Benjamini--Hochberg correction at a false discovery rate of $q<0.05$.
    }
    \label{fig:mlr}
\end{figure*}

\begin{figure*}[!htbp]
    \centering
    \includegraphics[width=0.85\textwidth]{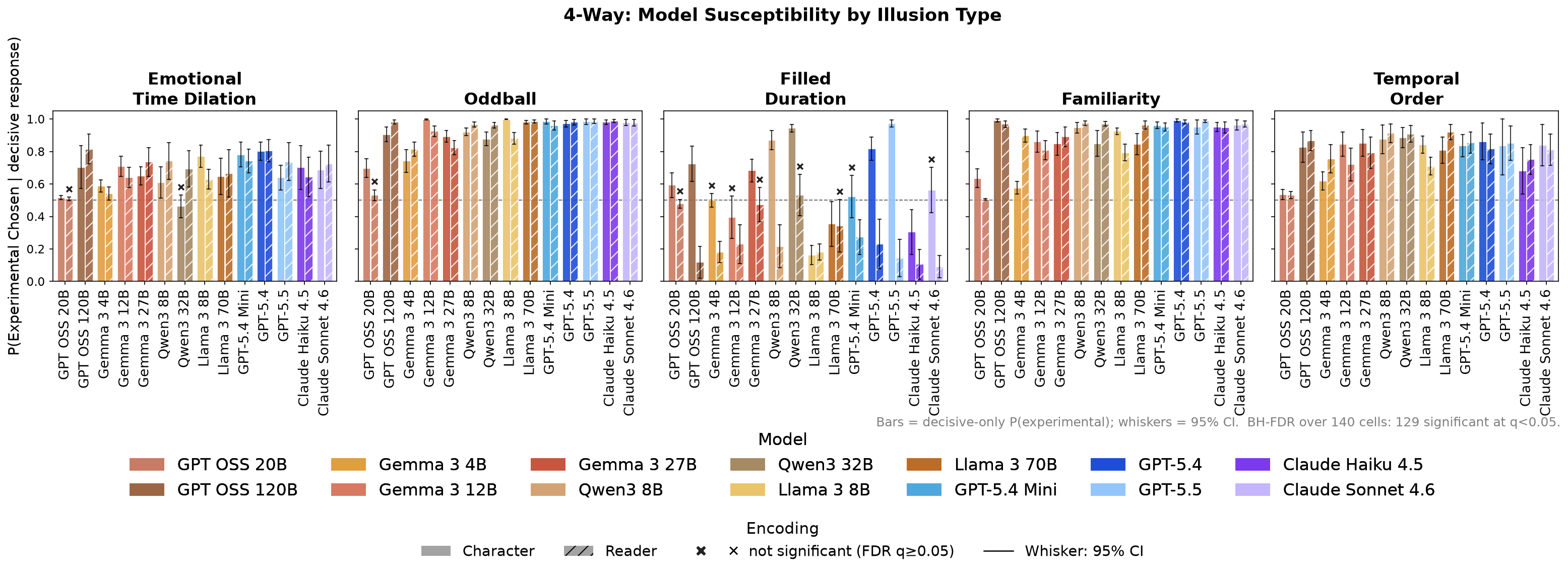}
    \caption{
    Model agreement across illusion types under 4-way evaluation. Bars represent the decisive-only proportion $p_{\text{exp}}$ of responses selecting the experimental scenario; whiskers are 95\% cluster-robust confidence intervals. Cells marked $\times$ are not significant under Benjamini--Hochberg correction at a false discovery rate of $q<0.05$.
    }
    \label{table:wilcoxon_effects}
\end{figure*}

Beyond illusion type and perspective, we analyze how additional factors, such as positional bias and prompt complexity (short vs.\ long), affect model behavior. Although we used Qwen3-32B to generate paraphrased options for the large-scale dataset, its choices do not appear to systematically differ from those of Qwen3-8B, though only Qwen3-32B's traces use research framing (Appendix~\ref{sec:trace_patterns}; compare the two Qwen3 models' per-model results in Figures~\ref{fig:mlr} and~\ref{table:wilcoxon_effects}). This suggests that the use of a Qwen model for data generation does not introduce a strong model bias in the observed results.

Models exhibit systematic differences in uncertainty: smaller open-source models show stronger positional bias. Models with greater uncertainty also demonstrate higher rates of non-committal responses. Across model families, gpt-oss models tend to avoid decisions under ambiguity, Qwen models more explicitly express uncertainty, and Gemma models typically force a selection. Prompt length has limited impact across most illusion types, with the exception of filled duration, where longer narratives strengthen the expected effect.

We observe substantial \textbf{positional bias}, particularly in the smaller open-source models. In the 4-way evaluation, the smallest model within each open-source family has the largest positional bias. Larger and proprietary models generally show smaller gaps, though Gemma 3 27B has a larger gap than Gemma 3 12B, and the two gpt-oss models are nearly equal under 2-way (Table~\ref{tab:positional_bias}). To mitigate these effects, we apply a debiasing strategy used in~\citet{li-etal-2020-unqovering} that symmetrizes results across option orderings. A paired McNemar test confirms substantial position bias for many (mainly open-source) models.

\begin{figure*}[!t]
    \centering
    \includegraphics[width=0.95\textwidth]{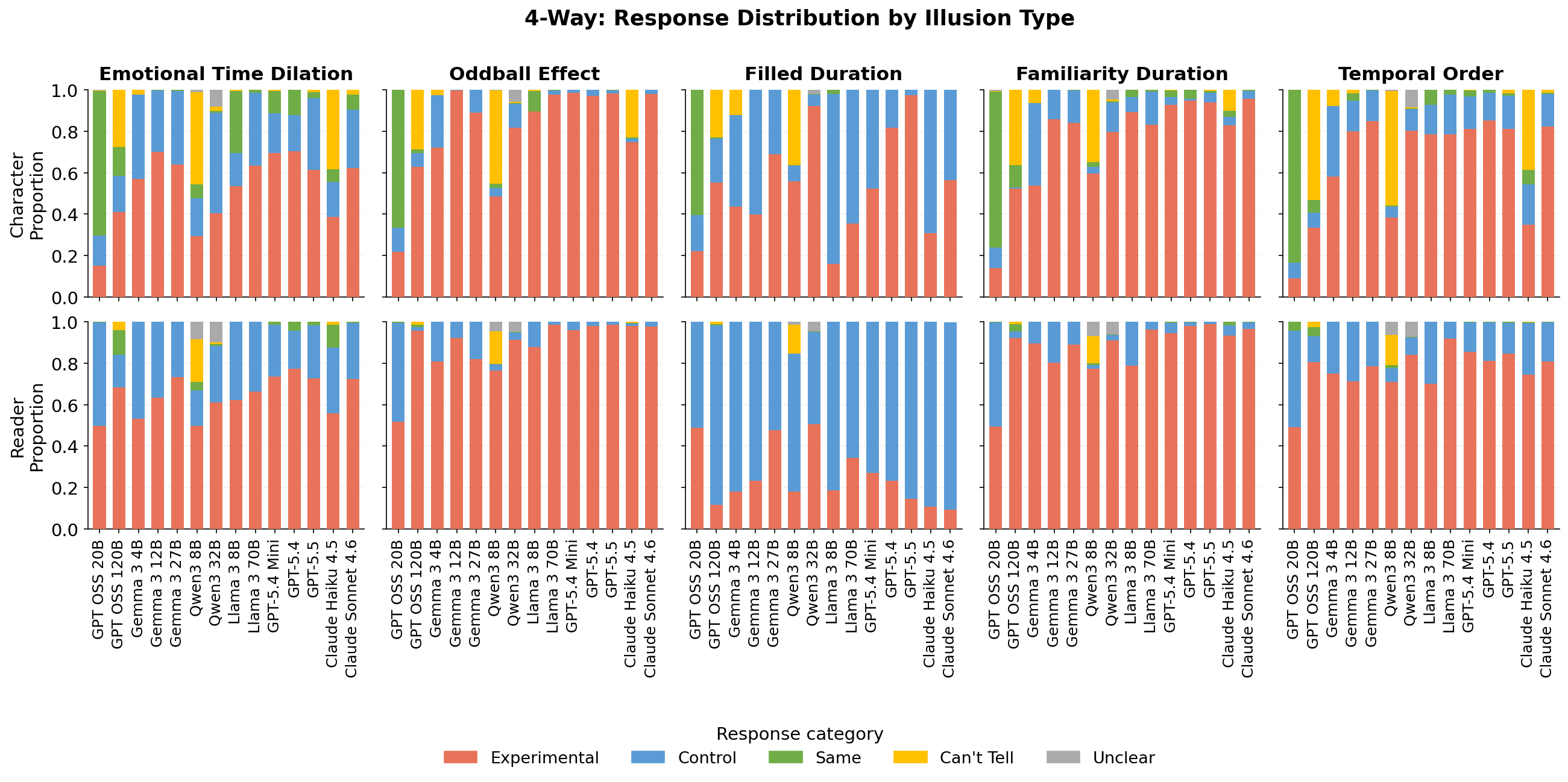}
    
    \caption{
    Model response distributions across illusion types and perspectives for 4-way evaluation.
    }
    \label{fig:stacked_response_rates}
\end{figure*}

Positional influence appears more pronounced in models when there is greater uncertainty among the leading candidate choices~\cite{pezeshkpour2024large}, demonstrated through \textbf{noncommittal answers}. Prior research indicates that introducing ``None of the Above'' (NA) as a correct option can result in performance decrements of 30–50\%, as models often struggle to reject all provided choices~\cite{tam2025none}. In our evaluation, smaller models exhibit both greater positional sensitivity (Table~\ref{tab:positional_bias}) and higher rates of non-committal responses (Figure~\ref{fig:stacked_response_rates}), suggesting that underlying uncertainty may modulate positional bias. Conversely, larger, more capable models demonstrate less dependence on option ordering and a lower propensity to abstain from a definitive choice. Taken together, these findings suggest that positional bias and non-committal behaviors may be linked to model uncertainty, potentially serving as heuristic strategies when a clear answer cannot be determined.

\begin{table}[t]
\centering
\scriptsize
\setlength{\tabcolsep}{3pt}
\renewcommand{\arraystretch}{0.95}
\begin{tabular}{lccc}
\toprule
\textbf{Model} & \textbf{Exp 1st (\%)} & \textbf{Exp 2nd (\%)} & \textbf{Gap (\%)} \\
\midrule
\multicolumn{4}{c}{\em 4-way evaluation}\\
gpt-oss-20b    & 68.8 & 56.1 & 12.7 \\
gpt-oss-120b   & 76.6 & 73.1 & 3.5 \\
Gemma 3 4B     & 85.1 & 35.2 & 49.9 \\
Gemma 3 12B    & 84.9 & 51.9 & 33.0 \\
Gemma 3 27B    & 93.7 & 55.7 & 38.0 \\
Qwen3-8B       & 85.5 & 68.4 & 17.1 \\
Qwen3-32B      & 84.2 & 74.3 & 9.9 \\
Llama 3 8B     & 76.2 & 53.9 & 22.3 \\
Llama 3 70B    & 79.6 & 64.6 & 14.9 \\
GPT-5.4 mini   & 72.1 & 80.2 & -8.2 \\
GPT-5.4        & 80.6 & 80.7 & -0.1 \\
GPT-5.5        & 77.8 & 80.2 & -2.4 \\
Claude Haiku 4.5 & 73.9 & 58.6 & 15.3 \\
Claude Sonnet 4.6 & 72.6 & 74.3 & -1.7 \\
\midrule
\multicolumn{4}{c}{\em 2-way evaluation}\\
gpt-oss-20b       & 76.71 & 71.24 & 5.47 \\
gpt-oss-120b      & 74.08 & 68.57 & 5.51 \\
Gemma 3 4B        & 88.28 & 36.71 & 51.56 \\
Gemma 3 12B       & 86.61 & 53.75 & 32.86 \\
Gemma 3 27B       & 96.02 & 45.96 & 50.06 \\
Qwen3-8B          & 84.96 & 66.66 & 18.30 \\
Qwen3-32B         & 85.80 & 75.69 & 10.11 \\
Llama 3 8B        & 84.13 & 45.34 & 38.79 \\
Llama 3 70B       & 84.66 & 70.63 & 14.04 \\
GPT-5.4 mini      & 72.99 & 79.81 & -6.83 \\
GPT-5.4           & 79.43 & 82.59 & -3.16 \\
GPT-5.5           & 77.12 & 80.96 & -3.85 \\
Claude Haiku 4.5  & 76.05 & 65.12 & 10.93 \\
Claude Sonnet 4.6 & 73.14 & 78.25 & -5.12 \\
\bottomrule
\end{tabular}

\caption{
We measure positional bias as the difference in selecting the experimental option when it appears first versus second, reported for the 4-way (top block) and 2-way (bottom block) evaluations.
}
\label{tab:positional_bias}
\end{table}

gpt-oss-20b, especially, and the gpt-oss family as a whole, show a strong tendency toward non-committal responses (Figure~\ref{fig:stacked_response_rates}), with up to 85\% of answers falling into the \emph{same} category at times. In contrast, Qwen models are far less likely to avoid a decision, staying around $\sim$10\% non-selective overall and primarily selecting \emph{cannot tell} rather than defaulting to \emph{same}. In fact, Qwen is the only model family with a substantial number of responses where it explicitly refuses to select an answer. Gemma models, by comparison, almost always commit to one of the available options, rarely using either non-selective category. Models were generally more likely to select non-committal responses when evaluating events from the character's perspective. Smaller models were more likely to select a non-committal answer. Additionally, models tended to default to only one of the two non-committal options, highlighting differences in whether they prefer to force a selection or admit uncertainty. This reveals a clear behavioral distinction: gpt-oss models tend to resolve ambiguity by avoiding a decision, Qwen prefers to admit uncertainty, and Gemma consistently forces a choice.

\begin{figure*}[!t]
    \centering
    \includegraphics[width=0.95\textwidth]{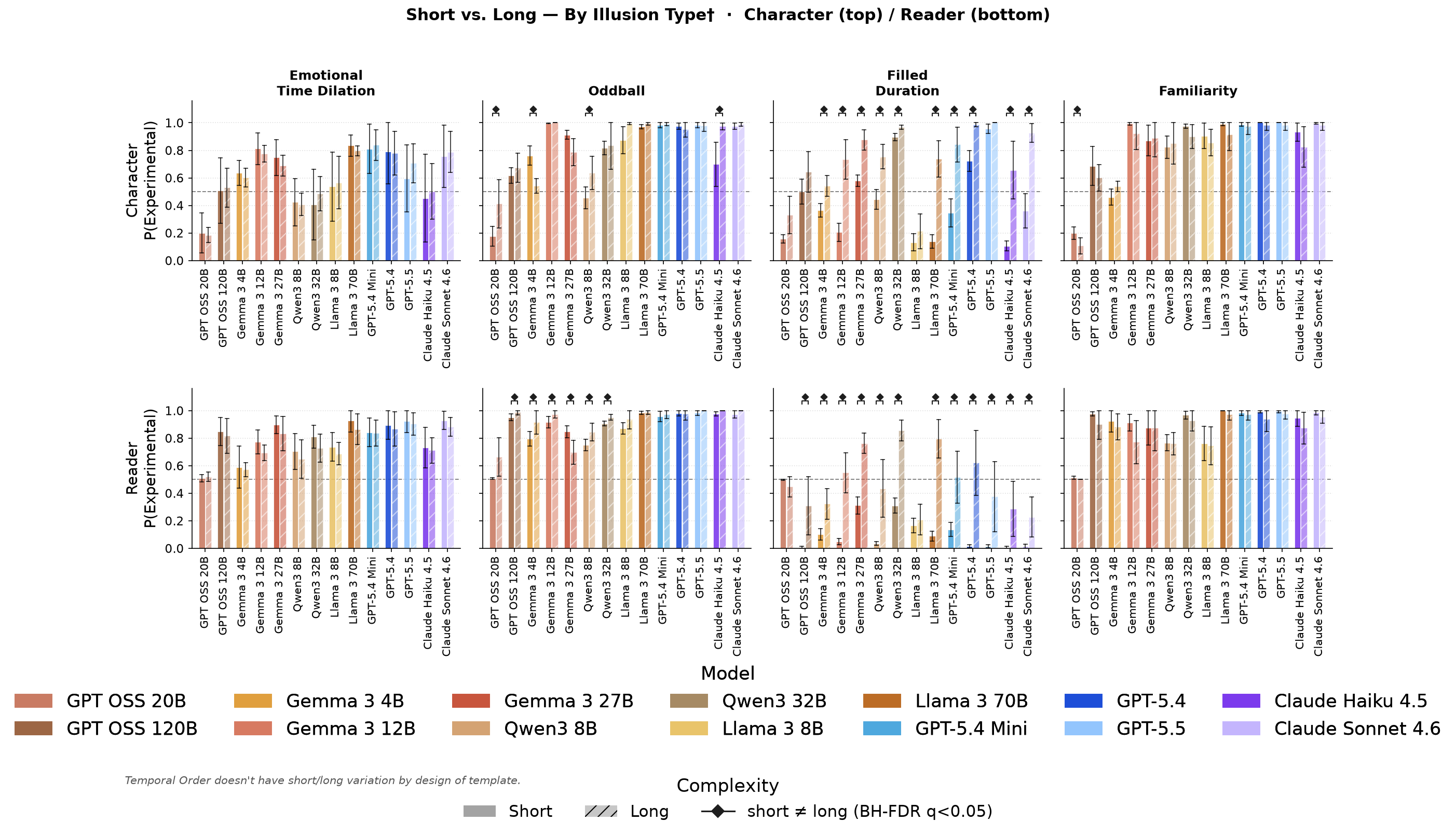}
    
    \caption{
    Comparison of model performance between short and long prompts across illusion types under 4-way evaluation. While certain models and illusion types show small differences, the overall pattern of effects remains largely consistent, indicating that prompt length has limited impact on the underlying signal captured by the model.
    }
    \label{fig:short_long_4way}
\end{figure*}

Across illusion types, \textbf{prompt length} has a relatively limited and inconsistent effect on model behavior (Figure~\ref{fig:short_long_4way}) across most illusion types. In most cases, both short and long prompts preserve the same overall pattern of illusion strength, suggesting that prompt length does not substantially alter the design.

One exception is the Filled Duration effect, in which longer narratives are more likely to yield selections aligned with the experimental option. This is consistent with the nature of the illusion: longer textual descriptions have more context, filling a small duration of time, thereby strengthening the effect.
In reverse, as the character, longer descriptions or more items may seem to make the duration seem longer. 
As a result, increased narrative length may directly impact perceived duration. Overall, while prompt length can modulate specific illusion types, it does not substantially change the relative ordering or general pattern of effects across conditions. This suggests that the length-matching condition, although not exact, was effective enough for the other illusions.

\textbf{Long items are under-represented after filtering.} The template library contains 122 templates (62 long, 60 short). Paraphrases of long narratives more often differed in word count from their source by more than seven words, so the paraphrase filter (\S~\ref{sec:benchmark_design}) removed long items far more often than short ones. All items from seven long templates were removed, leaving 115 templates in the full dataset. Templates were sampled uniformly, so we can estimate the pre-filter counts: the filter kept about 95\% of short items but only about 39\% of long ones (Table~\ref{tab:length_retention}). Long items make up 30\% of the full dataset (2,000 of 6,684) but 49\% of the gold set (123 of 250), which was curated by hand and not filtered. Short-vs.-long comparisons on the full dataset therefore rest on few long items for Oddball (214) and Temporal Order (175), and the surviving long items may not be representative of their templates. The gold set is balanced by length, so it gives the cleaner short-vs.-long comparison.

\begin{table}[ht]
\centering
\small
\setlength{\tabcolsep}{4pt}
\begin{tabular}{lrrrr}
\toprule
& \multicolumn{2}{c}{\textbf{Short}} & \multicolumn{2}{c}{\textbf{Long}} \\
\textbf{Illusion Type} & \textbf{Kept} & \textbf{\%} & \textbf{Kept} & \textbf{\%} \\
\midrule
emotional time dilation & 904 & 99 & 726 & 67 \\
oddball effect & 994 & 99 & 214 & 21 \\
filled duration effect & 1005 & 100 & 558 & 56 \\
familiarity duration effect & 885 & 92 & 327 & 32 \\
temporal order & 896 & 86 & 175 & 18 \\
\midrule
\textbf{Total} & \textbf{4684} & \textbf{95} & \textbf{2000} & \textbf{39} \\
\bottomrule
\end{tabular}
\caption{Short and long items kept after paraphrase filtering. Percentages are relative to the expected pre-filter count under uniform template sampling (2,000 items per illusion); pre-filter counts were not stored. Templates with no remaining items: one long Emotional Time Dilation, two long Oddball and four long Temporal Order templates.}
\label{tab:length_retention}
\end{table}

\subsection{Confound Analysis and Cue-Neutralization Details}
\label{sec:confound_details}
\paragraph{Lexical and semantic confound analysis.} We measured how the experimental and control passages differ on the surface and asked whether those differences explain the choices. For each of the 250 gold items, we scored both passages on 15 measures in 10 feature groups (Length, Word frequency, Concreteness, Sentiment, Affect, Predictability, Prompt overlap, Temporal language, Cue words, and Structure) and then took the difference between them (experimental minus control). Refer to Table~\ref{tab:confound-features} for feature details. We used a logistic regression to predict whether the experimental passage was chosen from these 15 measures in 10 feature groups, also accounting for illusion type, perspective (character's or reader's) and which passage was shown first.

\paragraph{Cue-neutralization test.} Excluding the items that did not preserve the original events and order does not change the results. Rewrites were also checked automatically (cue-word list and an experimental/control length ratio between 0.8 and 1.25, with retries); the 25 of 250 cue-neutralized rewrites that failed were left out of the evaluation. Because Gemma 3 27B is also an evaluated model, we re-ran the analysis without it, and results differ by about 0.03. The specific removed word depends on the illusion, such as emotional and arousal for emotional time dilation and novelty and familiarity for the familiarity-duration effect. Both the experimental and control passages were edited in the same way. These two alterations compare the effect of removing cues with the effect of just rewording. We evaluated the models on the original, paraphrase, and neutralized versions with identical prompts (Table~\ref{tab:cue-example}).

\begin{table*}[t]
\centering
\footnotesize
\setlength{\tabcolsep}{4pt}
\renewcommand{\arraystretch}{1.05}

\begin{tabularx}{\textwidth}{@{}l l X@{}}
\toprule
\textbf{Feature group} & \textbf{Measure} & \textbf{Method} \\
\midrule
\multirow{2}{*}{Length}
  & Characters & Total number of characters in the passage \\
  & Tokens & Total number of GPT-2 tokens in the passage \\
Word frequency & Mean Zipf frequency & Mean word-level Zipf frequency from \texttt{wordfreq} \cite{robyn_speer_2022_7199437} \\
Concreteness & Mean concreteness & Mean concreteness rating across content words \cite{Brysbaert2014} \\
\multirow{2}{*}{Sentiment}
  & Sentiment & VADER compound sentiment score \cite{HuttoGilbert2014} \\
  & Emotional intensity & Absolute VADER compound score, $\lvert\text{compound}\rvert$ \\
\multirow{2}{*}{Affect}
  & Valence & Mean NRC-VAD valence across content words \cite{mohammad-2018-obtaining} \\
  & Arousal & Mean NRC-VAD arousal across content words \cite{mohammad-2018-obtaining} \\
Predictability & Mean surprisal & Mean token-level surprisal (bits) under GPT-2 Medium \cite{radford2019language} \\
Prompt overlap & Question overlap & Proportion of content words shared with the question prompt \\
Temporal language & Time words & Count of explicit temporal expressions (e.g., \textit{minutes}, \textit{while}, \textit{finally}) \\
Cue words & Psychology cue words & Count of words associated with the target illusion (e.g., \textit{unexpected}, \textit{familiar}, \textit{busy}) \\
\multirow{3}{*}{Events \& structure}
  & Verbs & Number of verbal predicates from the dependency parse \\
  & Clauses & Number of clauses from the dependency parse \\
  & Syntactic depth & Maximum depth of the dependency parse tree \\
\bottomrule
\end{tabularx}
\caption{The 15 measures in 10 feature groups used in the confound analysis. Each feature was computed for the experimental and control passage of every item, and the analysis uses the difference (experimental $-$ control). Parsing, part-of-speech tags and lemmas come from spaCy~\cite{honnibal2020spacy}.}
\label{tab:confound-features}

\vspace{1.5em}

\begin{tabularx}{\textwidth}{@{}l X X@{}}
\toprule
\textbf{Version} & \textbf{Experimental passage} & \textbf{Control passage} \\
\midrule
Original &
John checked safety inspection forms. During the sixth check, one had fainter ink \textbf{than usual}. &
John \textbf{repeatedly} checked safety inspection forms. Each interval was \textbf{consistent}. He finished them. \\
\addlinespace
Paraphrase &
John inspected the safety documents, and the last one showed a noticeably pale inscription. &
John checked the safety documents, and each one was completed with standard clarity. \\
\addlinespace
Cue-neutralized &
John checked safety inspection forms. During the sixth check, one had lighter ink. &
John checked safety inspection forms. Each check followed the last. He finished them. \\
\bottomrule
\end{tabularx}
\caption{Example of the two rewrites for one oddball item. Words in \textbf{bold} in the original are cue words that signal the manipulation (expectation and repetition terms); they are removed in the cue-neutralized version, which keeps the same character, events and event order. The paraphrase rewords the item without restricting cue words.}
\label{tab:cue-example}
\end{table*}

\begin{figure*}[!htbp]
    \centering
    \begin{subfigure}[t]{0.45\textwidth}
        \centering
        \includegraphics[width=\linewidth]{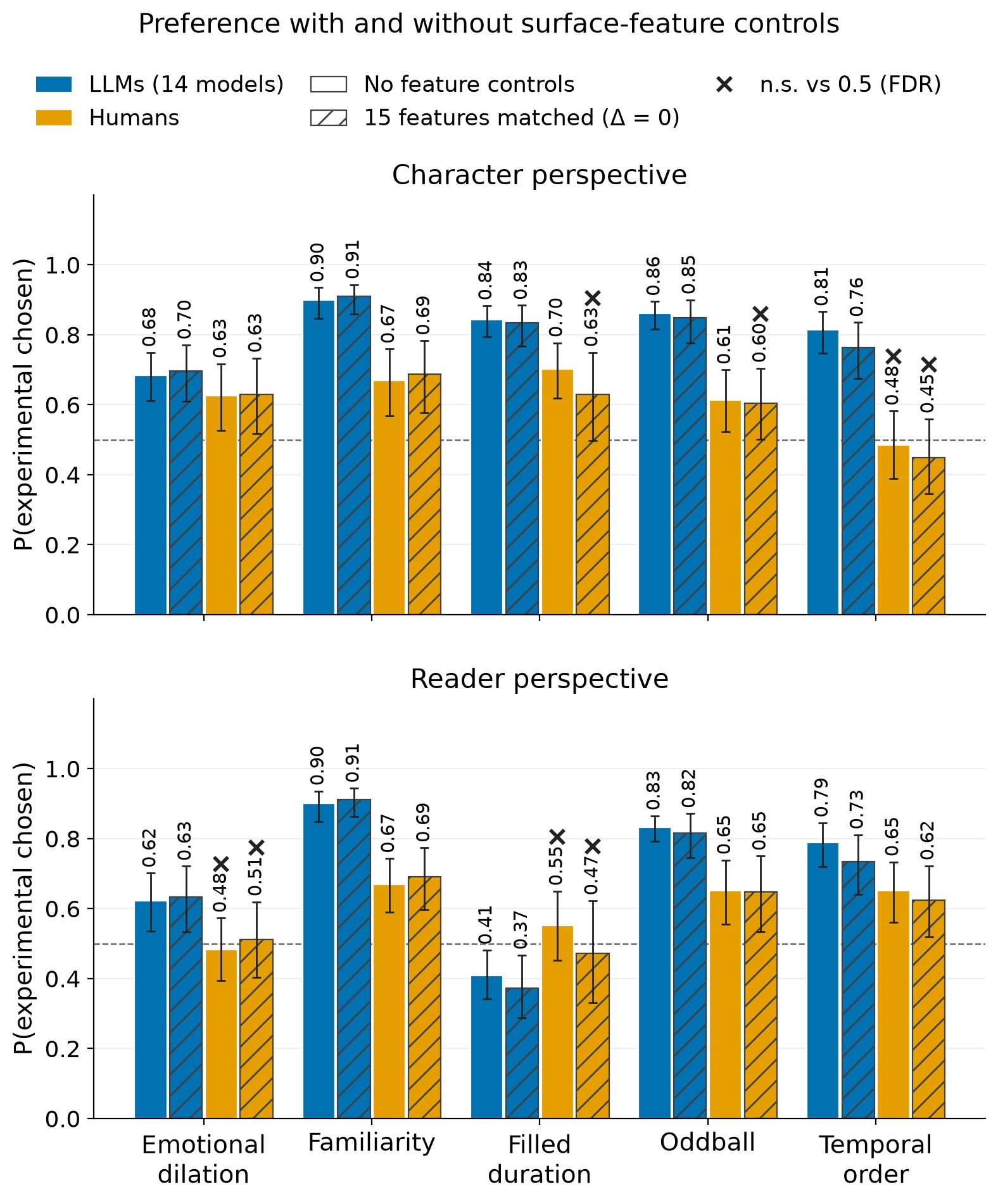}
        \caption{}
        \label{fig:residual_preference}
    \end{subfigure}
    \hfill
    \begin{subfigure}[t]{0.45\textwidth}
        \centering
        \includegraphics[width=\linewidth]{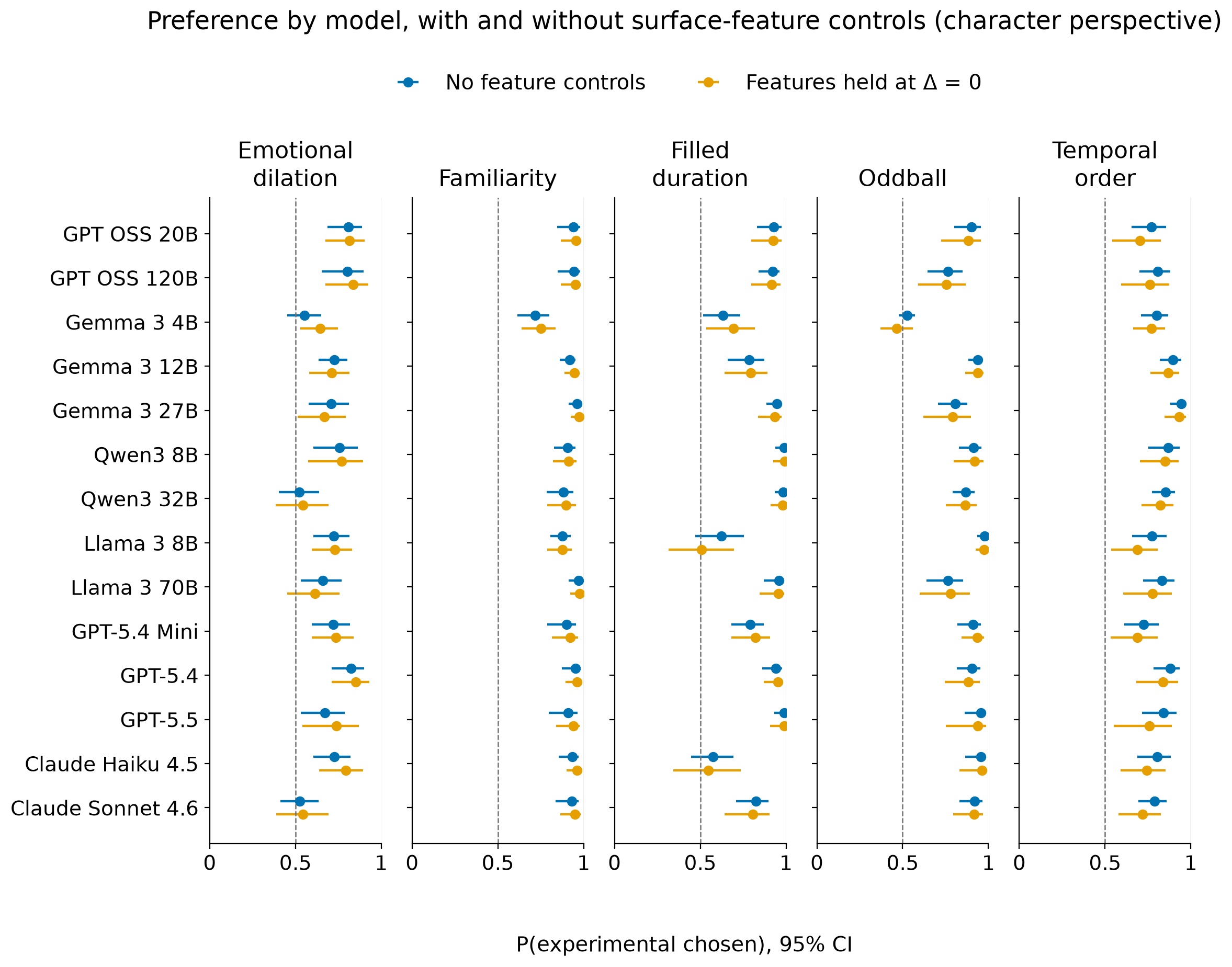}
        \caption{}
        \label{fig:residual_by_model}
    \end{subfigure}

    \caption{Experimental-choice preferences on the gold set before and after controlling for 15 lexical and semantic surface features: (a)~by perspective, for LLMs and for humans (mixed-effects model, as in Figure~\ref{fig:human_illusions_results}); (b)~by model, character's perspective. The controls widen the human intervals, so some cells (oddball and filled duration from the character’s perspective) are no longer significant; the oddball estimate barely moves, while filled duration drops from 0.70 to 0.63.}
    \label{fig:residual_analysis}
\end{figure*}

\subsection{Differences across Subtypes within Illusion Types}
\label{sec:subtypes}

\begin{figure}[!htbp]
    \centering
    \includegraphics[width=\columnwidth]{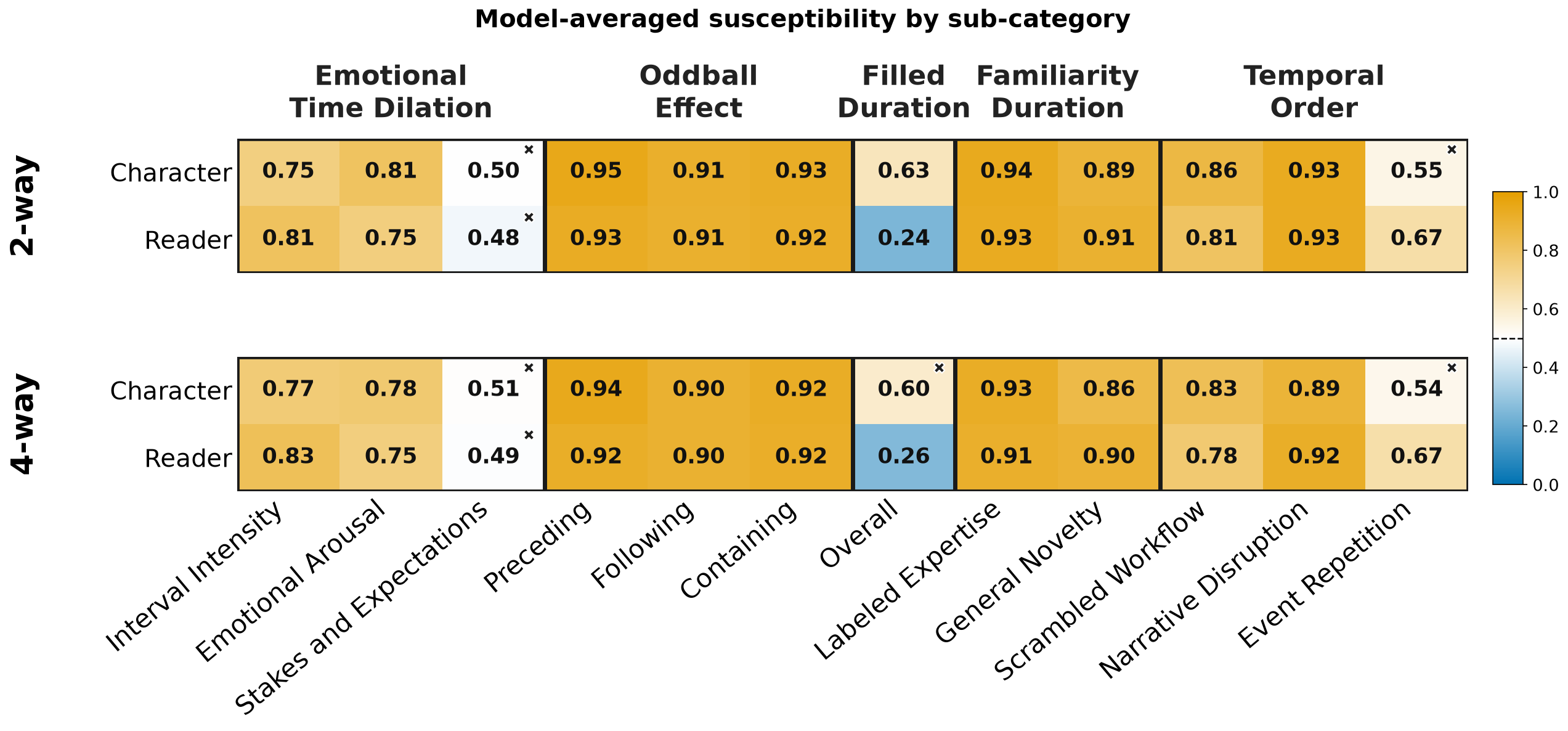}

    \caption{
    Each value in the heatmap represents the P(Experimental Chosen) for a subcategory within each illusion type, computed separately for the character's and reader's perspectives. }
    \label{fig:mlr_subcategory_heatmap_avg}
\end{figure}

Within each illusion type, Figure~\ref{fig:mlr_subcategory_heatmap_avg}
shows the relative differences across its subcategories, illustrating how individual components contribute to the overall effect. We excluded non-decisive selection options for the 4-way evaluation.

Within Emotional Time Dilation, Interval Intensity and Emotional Arousal yield much higher values than Stakes-and-Expectations, which stays near chance. This contrast suggests that LLMs capture directly observable cues more reliably than subcategories that require inferring stakes from context. The strong and consistent performance across Oddball Effect subtypes, driven primarily by clear, structured temporal deviations, further supports this interpretation.

For Temporal Order, Narrative Disruption exhibits the strongest signal and almost no difference between the reader's and character's perspectives. This subtype retains a relatively strong signal, potentially because it follows a linear progression with a clear deviation. In contrast, other Temporal Order subcategories, such as those involving repetition or subtle ordering constraints, show more variability. These subtypes require higher-level contextual or social reasoning (e.g., determining whether repeated or reordered events are meaningful or irregular).

Overall, these results reinforce our previous observation that situations grounded in clear, structured temporal deviations tend to produce more consistent model behavior.

\begin{figure}[htbp]
    \centering
    \includegraphics[width=\columnwidth]{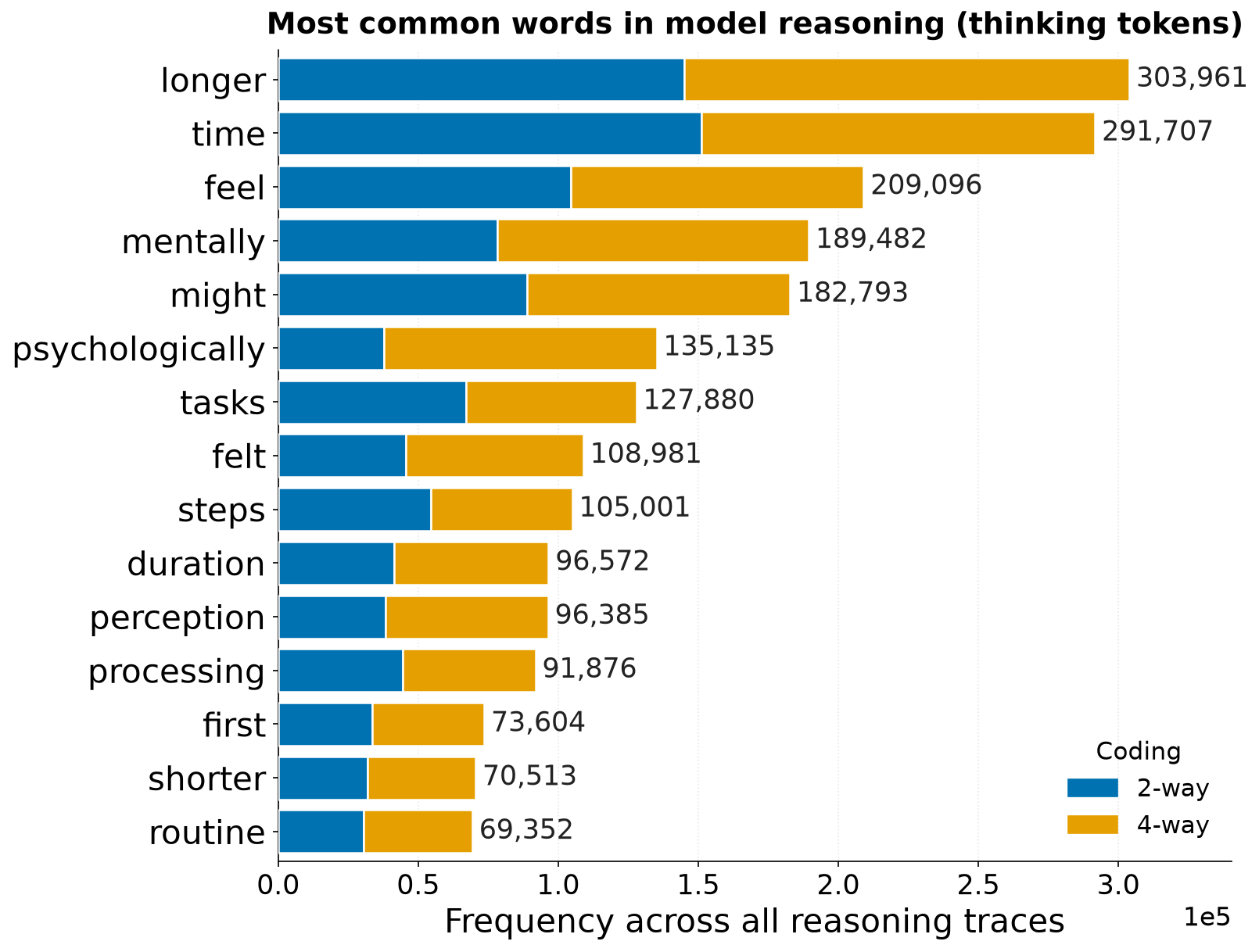}

    \caption{
    The most common words for all 5 illusion types split by 2-way and 4-way answer options within the thinking tokens of Qwen3-32B. Refer to Figure~\ref{fig:top_words} for a by illusion-type breakdown.
    \label{fig:thinking_tokens_agro}
    }
\end{figure}

\begin{figure*}[!htbp]
    \centering
    \includegraphics[width=0.85\textwidth]{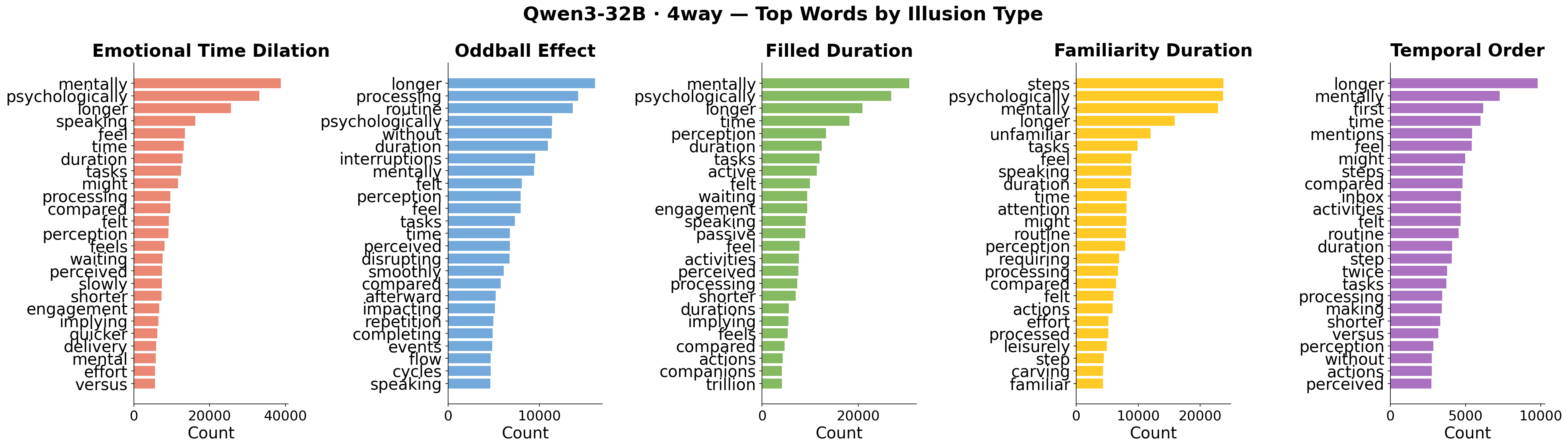}
    \caption{
    The most common words shown by models in their thinking tokens for each illusion category.
    }
    \label{fig:top_words}
\end{figure*}

\begin{figure*}[!htbp]
    \centering
    \includegraphics[width=0.95\textwidth]{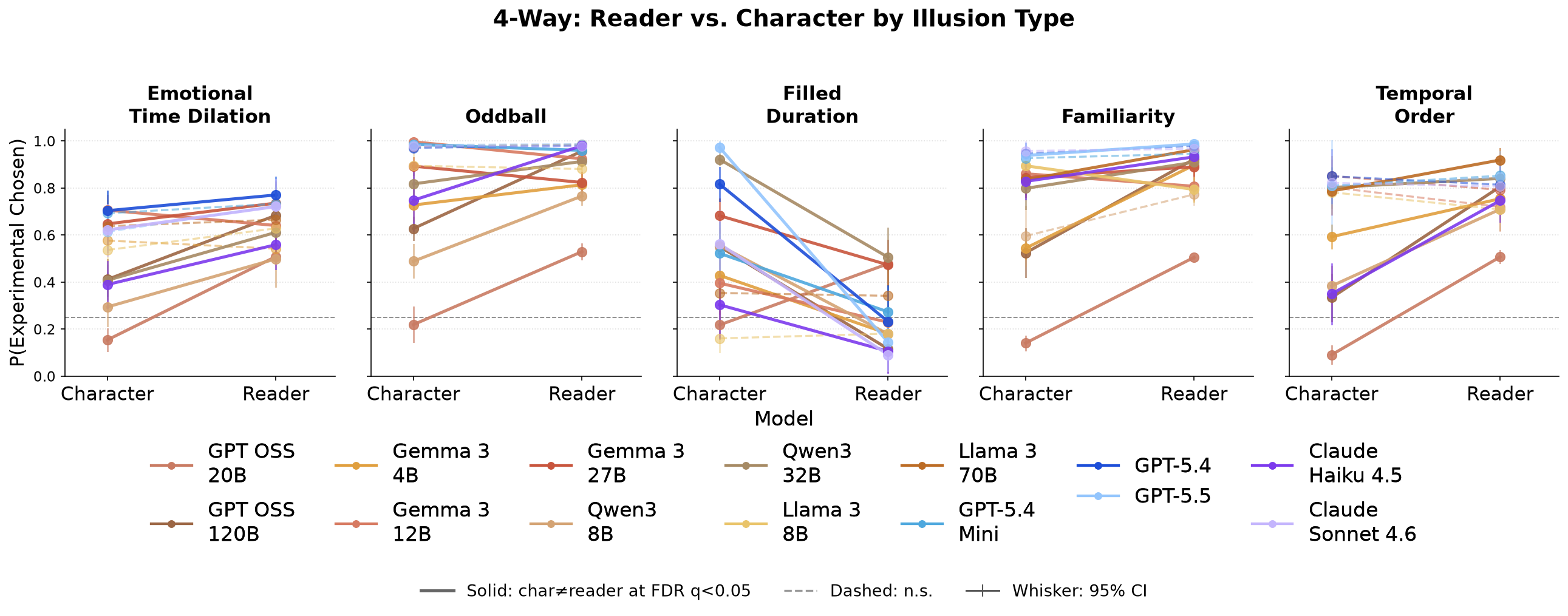}
    \caption{
    Comparison of the reader's and character's perspectives under 4-way evaluation across illusion types. Each line connects the same model evaluated under the two perspectives.
    }
    \label{table:reader_character_4way}
\end{figure*}

\begin{figure*}[!htbp]
    \centering
    \includegraphics[width=0.85\textwidth]{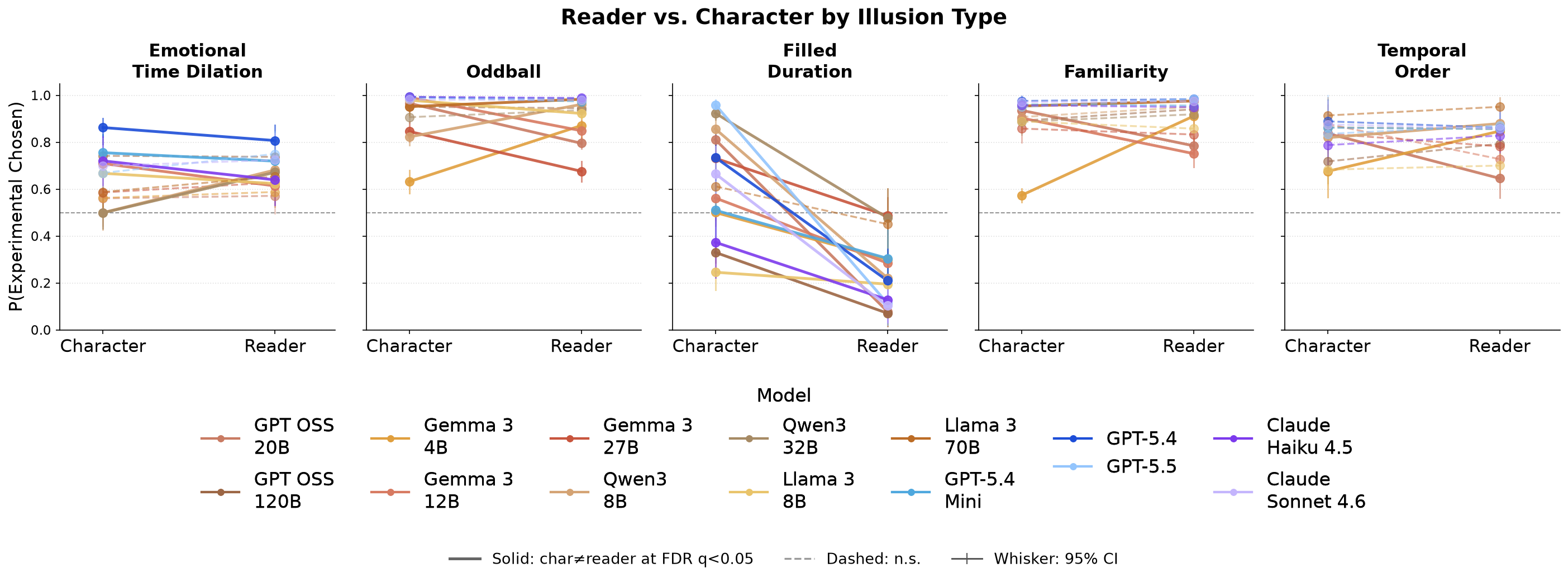}
    \caption{
    Comparison of the reader's and character's perspectives under 2-way evaluation across illusion types. Each line connects the same model evaluated under the two perspectives.
    }
    \label{fig:reader_character_2way}
\end{figure*}

\subsection{Research-Framing Patterns}
\label{sec:trace_patterns}
We analyzed Qwen3-32B because it is one of the largest evaluated models that exposes its thinking tokens. Its traces showed a recurring tendency to frame reasoning in terms of psychological phenomena. Word counts with NLTK~\cite{bird2009natural} showed that terms such as \textit{mentally} and \textit{psychologically} appeared across all five illusion categories (Figure~\ref{fig:thinking_tokens_agro}). We labeled a trace as research-framed if it contained at least one match for an explicit appeal to research or a named theory. Bare words such as \textit{psychologically} or \textit{cognitive} did not count, and none of the matches came from the passage text itself. Example matches include `cognitive load theory', `psychological studies suggest' and `In psychology, time perception'. Because research-framed and other traces choose the experimental option at about the same rate, we cannot tell whether the framing reflects recall of the illusion literature or generic task framing.

Research framing appeared in 386 of 1000 Qwen3-32B traces. Research-framed Qwen3-32B traces chose the experimental option at 80.7\% vs.\ 81.7\% for other traces ($-1.0$ pts, 95\% CI $-6.8$ to $+4.7$; $-2.0$ [$-7.7$, $+3.5$] adjusted).
A trace was flagged if it contained at least one case-insensitive regular-expression match from the list below.
\begin{itemize}
    \item \textit{psychological / psychology / scientific / empirical / neuroscience / cognitive science / time-perception} + \textit{research, study/studies, literature, finding(s), evidence, theory/theories, experiment(s), principle(s)}
    \item \textit{research / studies / experiments / findings / evidence} + \textit{show(s/n), suggest(s), indicate(s), found/find(s), demonstrate(s/d), say(s)}, with an optional \textit{has/have} in between
    \item \textit{some / many / several / other} + \textit{studies / research(ers) / psychologists}
    \item \textit{researcher(s)}
    \item \textit{in psychology}
    \item \textit{according to} + \textit{psychology / research / studies / science / theory}
    \item Named theories: \textit{cognitive load, attentional gate, scalar timing, contextual change, storage size, information-processing, (event) segmentation, internal clock, pacemaker} + \textit{theory / model / hypothesis}
    \item Named effects: \textit{oddball, filled-duration, holiday paradox, Vierordt's, kappa, tau} + \textit{effect / illusion / paradox / law}
    \item \textit{known as (the)} \ldots\ \textit{effect / illusion / paradox}
\end{itemize}
Matches were mostly `cognitive load theory' (52\%), `psychological studies' (22\%), `in psychology' (15\%), `psychological principle(s)' (6\%), and `studies suggest/show' (3\%).

\subsection{Additional Human Analysis Visualizations}
\label{sec:human_vis}

\begin{figure*}[!htbp]
    \centering
    \includegraphics[width=0.85\textwidth]{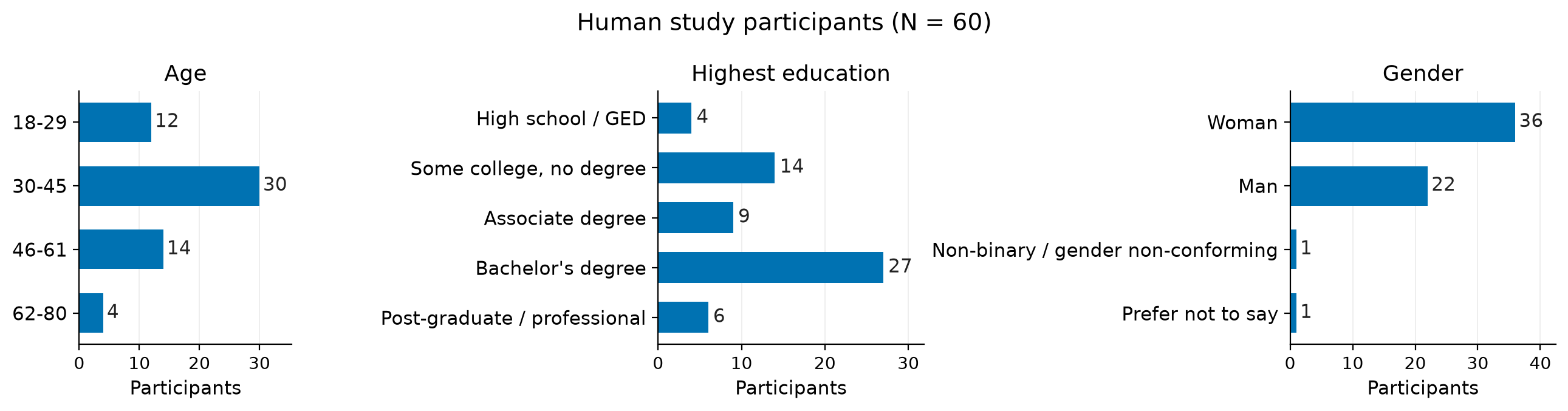}
    \caption{
    Number of human-evaluation participants ($N=60$) by age range, education and gender.
    }
    \label{fig:human_demographics}
\end{figure*}

Eligibility criteria, enforced using Prolific’s prescreening tools, required participants to be at least 18 years old, reside in the United States, identify English as their primary language, have completed at least one prior study on Prolific, and maintain an approval rate of 90\% or higher. Data collected through Prolific were fully anonymous. We provided an optional demographic questionnaire at the end of the survey (age range, highest education level and gender), shown in Figure~\ref{fig:human_demographics}. We will release only anonymized responses.

\begin{figure*}[!htbp]
    \centering
    \includegraphics[width=0.78\textwidth]{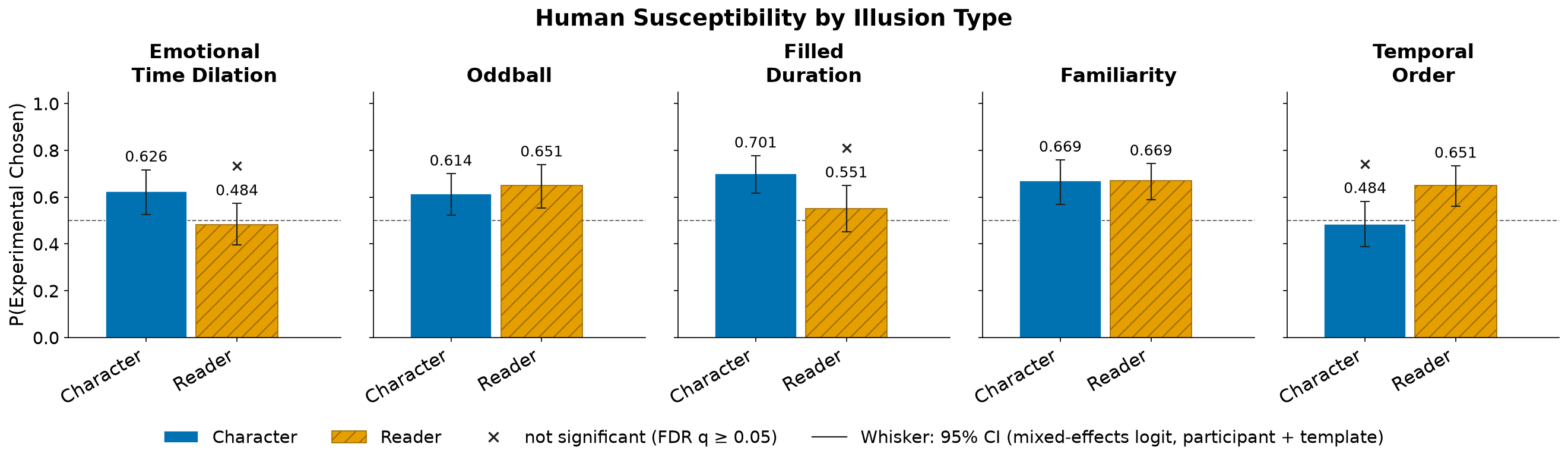}
    \caption{
    Human evaluation across illusion types following a 2-way evaluation with positional bias correction.
    }
    \label{fig:human_illusions_results}
\end{figure*}

\section{Use of AI Assistants}
\label{appendix:ai_use}

We used AI assistants to help write data-analysis code and to edit the paper's writing for clarity. These artifacts were all reviewed, tested, and edited by the authors. LLMs used for dataset construction (GPT-4o, Qwen3-32B, Gemma 3 27B) are described in \S~\ref{sec:dataset_creation} and \S~\ref{sec:llm_eval} and Appendix~\ref{sec:confound_details}.

\end{document}